\documentclass[9pt,twocolumn,twoside]{pnas-new}

\templatetype{pnasresearcharticle} 

\usepackage{graphicx}
\usepackage{subfigure}
\usepackage{booktabs} 
\usepackage{amsmath,amsthm,amsfonts,thmtools}
\usepackage{amssymb}
\usepackage{multirow}
\usepackage{xcolor}
\usepackage{natbib}
\usepackage{siunitx}
\usepackage{lmodern}
\usepackage{comment}
\usepackage{tabularx}
\usepackage{soul}
\usepackage{algorithm}
\usepackage{algpseudocode}
\usepackage{euscript}

\renewcommand{\dates}[1]{}
\newcommand{\bx}{{\bf x}}
\newcommand{\bX}{{\bf X}}

\newcommand{\bb}{{\bf b}}

\newcommand{\bA}{{\bf A}}

\newcommand{\bY}{{\bf Y}}

\newcommand{\cG}{\mathcal{G}}
\newcommand{\cP}{\mathcal{P}}
\newcommand{\cN}{\mathcal{N}}
\newcommand{\cV}{\mathcal{V}}
\newcommand{\cE}{\EuScript{E}}

\newcommand{\R}{{\mathbb R}}

\newcommand{\cL}{{\mathcal L}}

\newcommand{\fref}[1] {Fig.~\ref{#1}}
\newcommand{\Tref}[1]{Table~\ref{#1}}

\newcommand{\rev}[1]{\textcolor{black}{#1}}
\newcommand{\secrev}[1]{\textcolor{black}{#1}}

\begin{document}

\title{Message-Passing Monte Carlo: Generating low-discrepancy point sets via Graph Neural Networks}

\author[a,1]{T. Konstantin Rusch}
\author[b]{Nathan Kirk}
\author[c]{Michael M. Bronstein}
\author[b]{Christiane Lemieux}
\author[a]{Daniela Rus}

\affil[a]{Massachusetts Institute of Technology (MIT)}
\affil[b]{University of Waterloo}
\affil[c]{University of Oxford}

\leadauthor{Rusch}

\significancestatement{
This article introduces \emph{Message-Passing Monte Carlo (MPMC)}, the first machine learning approach for generating low-discrepancy point sets which are essential for efficiently filling space in a uniform manner, and thus play a central role in many problems in science and engineering. To accomplish this, MPMC utilizes tools from Geometric Deep Learning, specifically by employing Graph Neural Networks. \rev{MPMC can be extended to a higher-dimensional case which further allows for generating custom-made points.} Finally, MPMC point sets significantly outperform previous methods, achieving near-optimal discrepancy in practice \secrev{for low dimension and small number of points, i.e., for which the optimal discrepancy can be determined}. This advancement holds promise for enhancing efficiency in fields like \rev{scientific computing}, computer vision, machine learning, and simulation.
}

\correspondingauthor{\textsuperscript{1}To whom correspondence should be addressed. E-mail: tkrusch@mit.edu}


\begin{abstract}
Discrepancy is a well-known measure for the irregularity of the distribution of a point set. Point sets with small discrepancy are called low-discrepancy and are known to efficiently fill the space in a uniform manner. Low-discrepancy points play a central role in many problems in science and engineering, including numerical integration, computer vision, machine perception, computer graphics, machine learning, and simulation. In this work, we present the first machine learning approach to generate a new class of low-discrepancy point sets named \textit{Message-Passing Monte Carlo (MPMC)} points. Motivated by the geometric nature of generating low-discrepancy point sets, we leverage tools from Geometric Deep Learning and base our model on Graph Neural Networks. We further provide an extension of our framework to higher dimensions, which flexibly allows the generation of custom-made points that emphasize the uniformity in specific dimensions that are primarily important for the particular problem at hand. Finally, we demonstrate that our proposed model achieves state-of-the-art performance superior to previous methods by a significant margin. \secrev{In fact, MPMC points are empirically shown to be either optimal or near-optimal with respect to the discrepancy for low dimension and small number of points, i.e., for which the optimal discrepancy can be determined. Code for generating MPMC points can be found at \href{https://github.com/tk-rusch/MPMC}{\textbf{https://github.com/tk-rusch/MPMC}}} 
\end{abstract}


\maketitle
\thispagestyle{firststyle}
\ifthenelse{\boolean{shortarticle}}{\ifthenelse{\boolean{singlecolumn}}{\abscontentformatted}{\abscontent}}{}

\firstpage[21]{2}

\dropcap{M}onte Carlo (MC) methods 
have been commonly used and are a popular choice for approximating and simulating complex real-world systems.
Known for their reliance on repeated random sampling, MC methods function well in problems involving optimization, numerical integration, and financial mathematics (particularly derivative pricing and risk management) via computer simulation. However, their convergence rate of $\mathcal{O}(N^{-1/2})$ in the number of samples $N$ means that achieving high precision 
with MC requires an impractically large number of samples for complex problems. To address this drawback, it is common to employ \textit{variance reduction techniques} such as importance sampling, stratified sampling, or control variates to obtain the same degree of accuracy with fewer samples (for details, see \cite{glasserman2004monte}, \cite{LEMIEUX2009} and references therein).

A particularly successful approach for 
convergence is called \textit{quasi-Monte Carlo} (QMC). QMC methods employ a deterministic point set, which replaces the purely random sampling with a sample whose points span the hypercube $[0,1]^d$ in a manner that is more uniform than what can be achieved with MC sampling. 
The fact that these point sets are constructed over $[0,1]^d$ is not overly restrictive as most, if not all, sampling algorithms used within the MC method take as input (pseudo)random numbers in $[0,1]$. 
The uniformity of these deterministic point sets (or indeed, any point set) can be captured by one of several of measures of irregularity of distribution, referred to by the umbrella term  \textit{discrepancy measures} \cite{DRMOTATICHY1997}. The more uniformly distributed the points are, the lower the discrepancy is; point sets possessing a small enough discrepancy value are called \textit{low-discrepancy}. In the classical setting, the {\em star-discrepancy}, widely regarded as the most important uniformity measure, of an $N-$element point set $\{\bX_i\}_{i=1}^N$ contained in $[0,1]^d$ represents the largest absolute difference between the volume of a test box and the proportion of points of $\{\bX_i\}_{i=1}^N$ that fall inside the test box, 
\begin{equation}
\label{eq:star_disc}
D^* \left(\{\bX_i\}_{i=1}^N \right) := \sup_{\mathbf{x} \in [0,1]^d} \left| \frac{\#\left( \{\bX_i\}_{i=1}^N \cap [0,\mathbf{x}) \right)}{N} - \mu([0,\mathbf{x})) \right|
\end{equation}
where $\#(\{\bX_i\}_{i=1}^N \cap [0,\bx))$ counts how many points of $\{\bX_i\}_{i=1}^N$ fall inside the box $[0,\bx) = \prod_{i=1}^d [0,x_i)$ for $\bx = (x_1,\ldots,x_d)\in [0,1]^d$, and $\mu(\cdot)$ denotes the usual Lebesgue measure. The discrepancy is closely related to worst-case integration error of a particular class of functions with the most well-known result being the Koksma-Hlawka inequality; see  \cite{KUIPNIED1974, HLAWKA1984}. Explicitly, given a point set $\{\bX_i\}_{i=1}^N$ contained in $[0,1]^d$, we have 
\begin{equation}\label{eq:KHinequality}
\left| \int_{[0,1]^d} f(\mathbf{x}) d\mathbf{x} - \frac{1}{N} \sum_{i=1}^N f(\bX_i) \right| \leq D^*(\{\bX_i\}_{i=1}^N) V(f)
\end{equation}
where $V(f)$ denotes the variation of the function $f$ in the sense of Hardy and Krause. This result illustrates that points with small discrepancy induce approximations with small errors. Thus in summary, it is of general interest to find $N-$point configurations with smallest discrepancy; see \cite{TEYTAUD2020, GALANTIJUNG1997, KELLER2022, mishra2021enhancing,longo2021higher} for examples of QMC implementation.

Given this context, our main goal is to present a machine learning framework that generates point sets with minimal discrepancy. Based on the geometric nature of this problem, we suggest to leverage graph-learning models from Geometric Deep Learning \cite{gdlbook} to achieve this. More concretely, we construct a computational graph based on nearest neighbors of the initial input points and process the encoded input points with a deep message-passing neural network, which is trained to minimize a closed-form solution of a specific discrepancy measure of its decoded and clamped outputs. We term the resulting low-discrepancy points \textbf{Message-Passing Monte Carlo (MPMC)} points. 
\rev{Previous methods either fail to achieve optimal discrepancy values or are computationally intractable, being limited to small dimensions ($d \leq 3$) and very small numbers of points ($N \leq 21$), which can still require weeks of computation. In contrast, MPMC can be trained in a few minutes to achieve near-optimal discrepancy for these cases. Moreover, we show that MPMC is not limited to small number of points in small dimensions but can efficiently generate $N>1000$ points in tens of dimensions.} 
This advancement represents a significant step forward in the development of highly efficient sampling methods, which are crucial for many applications in science and engineering. Concrete examples include problems in financial mathematics \cite{l2009quasi}, path and motion planning in robotics \cite{branicky2001quasi}, and enhanced training of neural scene rendering methods like Neural Radiance Fields (NeRFs) \cite{nerf}.

\paragraph{Main contributions.} In the subsequent sections, we will: 
\begin{itemize}
    \item introduce a new state-of-the-art machine learning model that generates low-discrepancy points.  
    To our knowledge, this is the first machine learning approach in this context.
    \item extend our framework to higher dimensions by minimizing the average discrepancy of randomly selected subsets of projections. This allows for generating custom-made points that emphasize specific dimensions that are primarily important for the particular problem at hand.
    \item provide an extensive empirical evaluation of our proposed MPMC point sets and demonstrate their superior performance over previous methods. 
\end{itemize}

\section{Background and previous work}\label{sec:background}

Our general goal in this paper is to provide a method for generating point sets with small discrepancy. In the following, we use the term {\em sequence} to refer to an infinite series of points, and {\em point set} for a finite one. 

A sequence of points $\{\bX_i\}_{i=1}^{\infty}$ contained in $[0,1]^d$ is called a low-discrepancy sequence if the star-discrepancy of the first $N$ points satisfies $D^*\left(\{\bX_i\}_{i=1}^N\right) = \mathcal{O}( (\log N)^{d} / N )$. A finite point set  $\{\bX_i\}_{i=1}^N$ is said to be of low discrepancy if its corresponding star-discrepancy $D^*\left(\{\bX_i\}_{i=1}^N\right)$ is ``small'' enough, which in practice means that a bound of the form
$c (\log N)^{d-1}/N$ can be established, for a given constant $c$ independent of $N$ (but possibly dependent on $d$).
Moving forward, for comparison purposes, we will truncate various known infinite low-discrepancy sequences resulting in a finite point set, which inherits the low-discrepancy property from the underlying infinite sequence.

\begin{figure}[t]
\vspace{-1em}
\begin{minipage}{.22\textwidth}
\includegraphics[width=1.\textwidth]{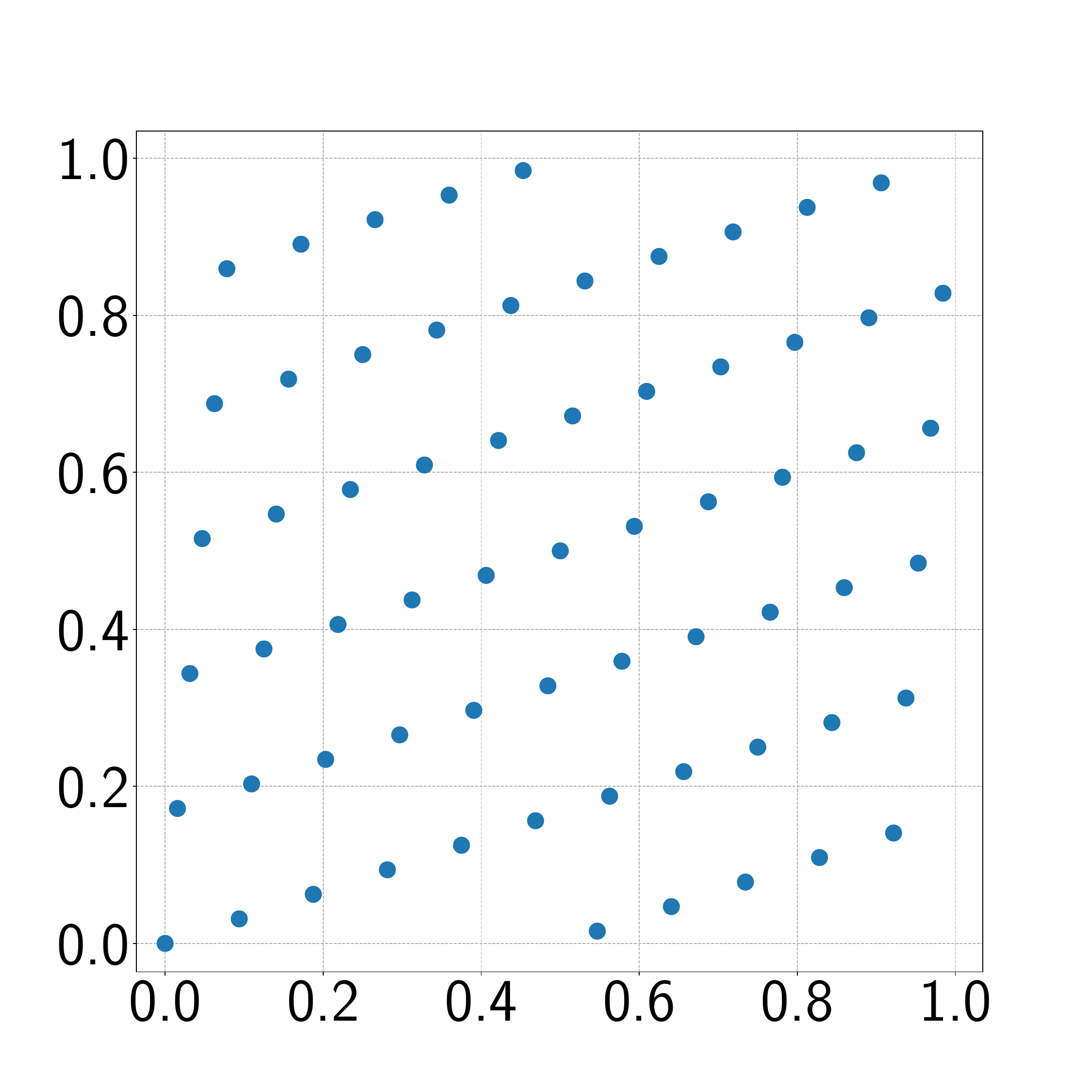}
\end{minipage}%
\hfill
\begin{minipage}{.22\textwidth}
\includegraphics[width=1.\textwidth]{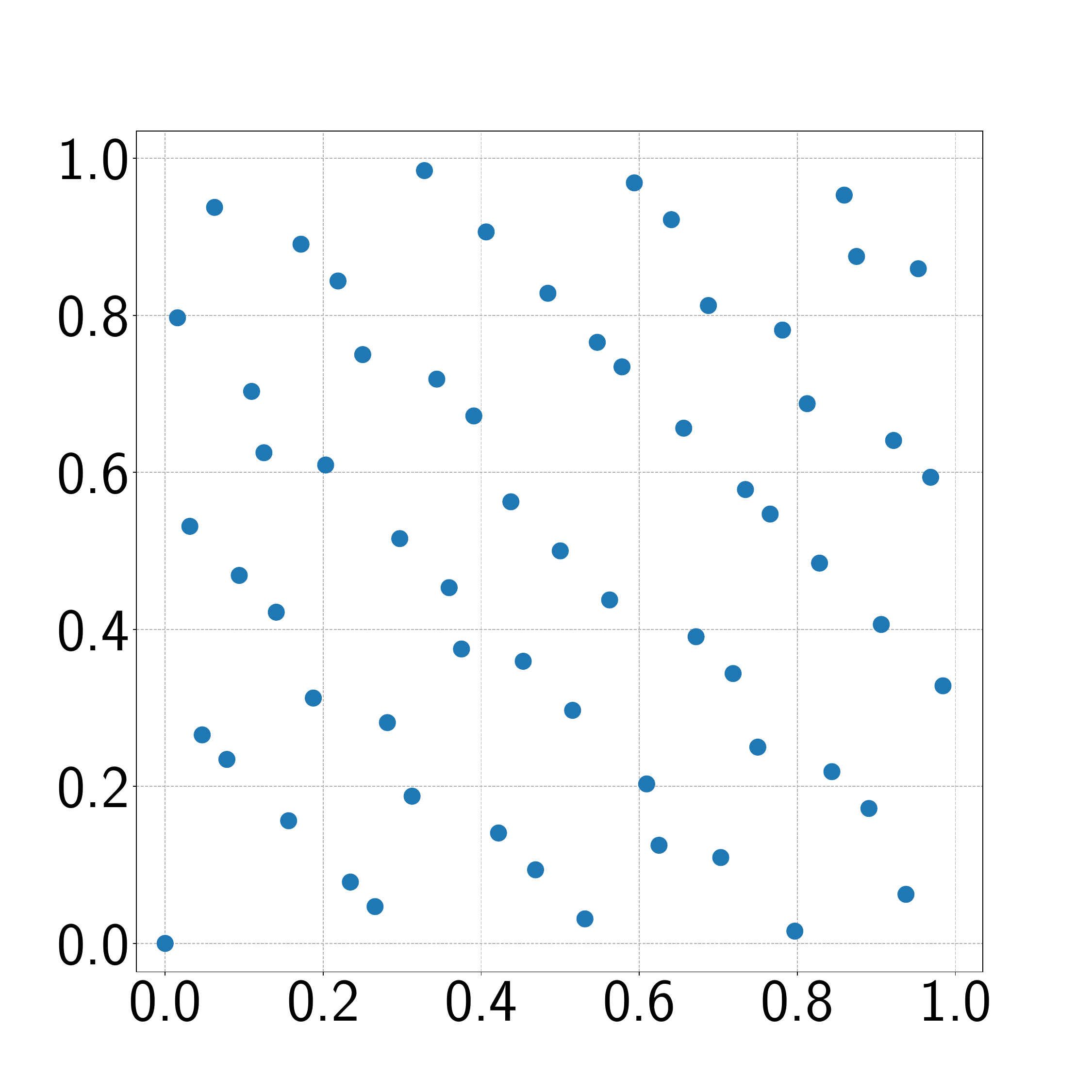}
\end{minipage}
\caption{Two different low-discrepancy point sets with $N=64$: Korobov lattice (left), and Sobol' (right).}
\label{fig:2ptsets}
\vspace{-1.em}
\end{figure}

Figure \ref{fig:2ptsets} illustrates two examples of low-discrepancy point sets. On the left-hand side we have a Korobov lattice \cite{KOROBOV1963}, which is an example of a lattice rule \cite{HABER1970, SLOANLATTICE1994, NUYENS2014, LATTICES2022}, and on the right-hand side, we see the first 64 points of the two-dimensional Sobol' sequence \cite{SOBOL1967}. This construction leverages a widely used building block for many low-discrepancy sequences known as the van der Corput sequence in base $b$  \cite{VDC1935}. It is also an example of what are modernly known as digital $(t,s)$-sequences---which also include the Faure sequences \cite{FAURE1982}---that were first laid out in \cite{NIED1987}, with a comprehensive overview provided in the subsequent monograph \cite{NIED1992}. Halton sequences \cite{HALTON1960} are another widely used type of low-discrepancy sequences that concatenate $d$ van der Corput sequences in different bases, usually taken as the first $d$ prime numbers.

More recently, there have been successful attempts to construct low-discrepancy point sets using more sophisticated means motivated by the lack of constructions adapted to specific $N$ and $d$. In \cite{DOERR2013}, new low-discrepancy point sets were suggested via the optimization of permutations applied to a Halton sequence. 
More recently, a method called \textit{subset selection} is formulated to choose from an $N-$element point set, the $k < N$ points which yield the smallest discrepancy. \rev{An exact selection algorithm was presented in \cite{CLEMENTDOERR2022}, while a swap-based heuristic approach was used in \cite{CLEMENTDOERR2024}.} Furthermore, a method to generate optimal star-discrepancy point sets for fixed $N$ and $d$ based on a non-linear programming approach was suggested in \cite{CLEMENTOPTIMAL2023}. However, this formulation of the problem presented huge computational burdens allowing optimal sets only to be found up to \rev{$21$} points in dimension two and $8$ points in dimension three.

\section{Method}
Let $1<d<+ \infty$ and $1\leq N < + \infty$ be fixed natural numbers. Our objective is to train a neural network to transform (random) input points $\{\bX_i\}_{i=1}^N$ into points $\{\hat{\bX}_i\}_{i=1}^N$ that reduce the star-discrepancy $D^*$ \eqref{eq:star_disc}, where $\bX_i, \hat{\bX}_i \in [0,1]^d$ for all $i$. 

In this work, we propose to leverage Graph Neural Networks (GNNs) \cite{sperduti1994encoding,goller1996learning,sperduti1997supervised,frasconi1998general,gori2005new,scarselli2008graph,bruna2013spectral,chebnet,gcn,MoNet} based on the message-passing framework to effectively learn such transformations. GNNs are a popular class of model architectures for learning on relational data, and have successfully been applied on a variety of different tasks, e.g., in computer science \cite{MoNet,eta_gat,ying2018graph}, 
and the natural sciences \cite{mpnn,gaudelet2021utilizing,shlomi2020graph} (see \cite{zhou,gdlbook} for additional applications). In particular, GNNs have successfully been used in the context of learning on point clouds, or generally learning on sets. This motivates the choice of GNNs in our setup, where specific transformations of geometric sets (i.e., set of input points in $[0,1]^d$) have to be learned.

\begin{figure}[ht]
\includegraphics[width=8.7cm]{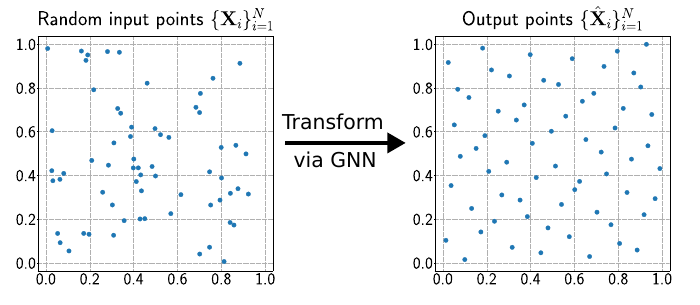}
\centering
\caption{Schematic drawing of our proposed approach to transform (random) input points $\{\bX_i\}_{i=1}^N$ into low-discrepancy points $\{\hat{\bX}_i\}_{i=1}^N$. Both the input and output point sets are actual instances of our proposed model, with $N=64$ and $d=2$ in this example.}
\label{fig:MPNN_approach_scheme}
\end{figure}

A schematic drawing of our approach can be seen in \fref{fig:MPNN_approach_scheme}, where we train a GNN model to transform $N=64$ random input points $\{\bX_i\}_{i=1}^N$ into low-discrepancy points $\{\hat{\bX}_i\}_{i=1}^N$.  

\subsection{Training set} 
\label{sec:inout_points}
Our approach can be classified as an unsupervised learning setup, where, in contrast to supervised learning, only input data is required without any labels. While it is intuitive to generate the set of input points randomly, we suggest several different approaches for constructing input data. \rev{First, using aforementioned uniform random sampled set of input points $\bX_i \sim \mathcal{U}([0,1]^d)$, for all points $i=1,\dots,N$. Second, using set of input points from available low-discrepancy constructions, such as Sobol', Halton, or a lattice rule. Third, using set of input points from randomly perturbed low-discrepancy points, i.e., \begin{equation}
\label{eq:randqmc}
\bX_i = \bY_i + \xi \hspace{1mm} (\text{mod } 1),
 \end{equation}
 where $\bY_i$ is generated by a known low-discrepancy set and $\xi$ is uniform randomly sampled from $[0,b]^d$, with $0<b\leq1$, for all points $i=1,\dots,N$.}

\subsection{Model architecture}
\rev{We start by constructing an undirected computational graph $\mathcal{G}=(\cV,\cE\subseteq \cV\times \cV)$, where $\cV$ denotes the set of unordered nodes corresponding to the input points $\{\bX_i\}_{i=1}^{N}$, and $\cE$ is the set of pair-wise connections between the nodes. In addition, each node $i \in \cV$ is equipped with a node feature set to the coordinates of an input point, i.e., set to $\bX_i \in \R^d$. We further denote the \emph{$1$-neighborhood} of a node $i \in \cV$ as \secrev{$\cN_i =\{j \in \cV : (i,j) \in \cE \}$}. Clearly, the set of all 1-neighborhoods induces the connectivity of the graph, i.e., the set of node-wise connections $\cE$. Hence, the only remaining part of the construction of the underlying computational graph $\cG$ is to define the local structure of the graph, i.e., defining $\cN_i$ for all nodes $i$. It is worth noting that in many GNN applications (e.g., network science, or life sciences) the computational graph structure is already given a-priori, either explicitly or implicitly. In contrast to that, our problem setup considers the connectivity of the underlying computational graph as an additional design choice. While there are many suitable choices, often balancing a global vs local connectivity structure, we suggest basing the local connectivity on the nearest neighbors of the node features, i.e., for a fixed radius $0<r\leq\sqrt{d}$,
\begin{equation}
\label{eq:radius_graph}
\cN_i = \{j \in \cV : \|\bX_i - \bX_j \|_2 \leq r\}.
\end{equation}
We choose this inherently local structure to guide the GNN training towards transforming input points into low-discrepancy points by mainly considering the positions of other near-by points (in the corresponding Euclidean space of the input point set).}  

\begin{figure*}[ht]
\includegraphics[width=13cm]
{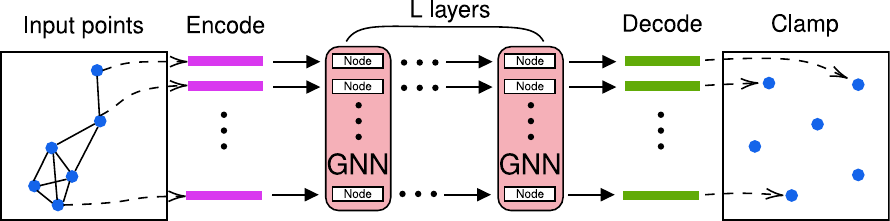}
\centering
\caption{Schematic of the proposed model to learn low-discrepancy points. First, (random) input points $\{\bX_i\}_{i=1}^N$ are encoded to a high dimensional representation. Second, the encoded representations are passed through a deep GNN \eqref{eq:MPMC_GNN}, where the underlying computational graph is constructed based on nearest neighbors using the positions of the initial input points. Finally, the node-wise output representations of the final GNN layer are decoded and clamped yielding new $d$-dimensional points $\{\hat{\bX}_i\}_{i=1}^N$ in $[0,1]^d$.}\label{fig:MPMC_model}
\end{figure*}

\rev{We can now define the main building block of our MPMC model, i.e., GNN layers based on the message-passing framework. Message-passing GNNs are a family of parametric functions defined through local updates of hidden node representations. More concretely, we iteratively update node features as,
\begin{equation}
\begin{aligned}
\label{eq:MPMC_GNN}
\bX_i^{l} = \phi^{l}\left(\bX_i^{l-1}, \rev{\sum\limits_{j \in \cN_i}} \psi^{l}(\bX_i^{l-1},\bX_j^{l-1})\right), \vspace{3mm} \text{for all } l=1,\dots,L,
\end{aligned}
\end{equation}
with $\bX_i^l \in \R^{m_l}$ for all nodes $i$. 
Moreover, we parameterize $\phi^l,\psi^l$ as ReLU-multilayer perceptrons (MLPs), i.e., MLPs using the element-wise $\text{ReLU}(x) = \max(0,x)$ activation function in-between layers.
We further encode the initial node features by an affine transformation $\bX_i^0 = \bA_{\text{enc}}\bX_i+\bb_{\text{enc}}$ for all $i=1,\dots,N$, with weight matrix $\bA_{\text{enc}}\in \R^{m_0 \times d}$ and bias $\bb_{\text{enc}}\in \R^{m_0}$. Finally, we decode the output of the final GNN layer by an affine transformation and smoothly clamp the decoded outputs back into $[0,1]^d$ by using the element-wise sigmoidal activation function, i.e., $\hat{\bX}_i = \sigma(\bA_{\text{dec}}\bX_i^L+\bb_{\text{dec}})$ for all $i=1,\dots,N$, with sigmoidal function $\sigma(x) = 1/(1+e^{-x})$, weight matrix $\bA_{\text{dec}}\in \R^{d \times m_L}$, and bias $\bb_{\text{dec}}\in \R^{d}$. Note that clamping is crucial, as otherwise the training objectives we introduce in the subsequent sections are ill-defined.} A schematic of the full model can be seen in \fref{fig:MPMC_model}.

\subsection{Training objective}
Our ultimate goal is to minimize the star-discrepancy $D^*$ \eqref{eq:star_disc}. However, $D^*$ cannot serve as the training objective, as (i) $D^*$ is computationally infeasible to calculate for high dimensions $d$ and large number of points $N$, (ii) the training objective should not only be computationally feasible but rather very efficient to compute, as it needs to be evaluated at every step of the training procedure (i.e., for every step of the gradient descent method) resulting in potentially thousands of evaluations to train only a single model, and (iii) the training objective needs to be \rev{sufficiently} differentiable \rev{(i.e., that can be handled by automatic differentiation packages such as \cite{PT_autograd})} in order to be used in the context of gradient-based learning. 
It turns out, we can derive a training objective resolving all three issues while simultaneously \rev{reducing} $D^*$ by leveraging previous work on the $\cL_p$-discrepancy,
\begin{equation}
\begin{aligned}
\label{eq:Lp_disc}
    &\cL_p(\{\bX_i\}_{i=1}^N) := \\&\left(\int_{[0,1]^d} \left| \frac{\#(\{\bX_i\}_{i=1}^N \cap [0,\bx))}{N} - \mu([0,\bx)) \right|^p d\bx \right)^{\frac{1}{p}}.
\end{aligned}
\end{equation}
Clearly, the star-discrepancy $D^*$ can be derived as a special case of \eqref{eq:Lp_disc} with $p=\infty$. Here, we focus on the case of $p=2$ as our training objective, since instead of computing the integral in \eqref{eq:Lp_disc}, we can leverage its closed-form expression, known as Warnock’s formula \cite{warnock1972computational},
\begin{equation}
\begin{aligned}
\label{eq:warnock}
\cL^2_2(\{\bX_i\}_{i=1}^N) &= \frac{1}{3^d} - \frac{2}{N} 
\sum_{i=0}^{N-1} \prod_{k=0}^d \frac{1-\bX_{i,k}^2}{2} \\&+ \frac{1}{N^2} \sum_{i,j=0}^{N-1} \prod_{k=0}^d 1-\max(\bX_{i,k},\bX_{j,k}),
\end{aligned}
\end{equation}
where $\bX_{i,k}$ is the $k$-th entry of $\bX_i$. This enables a very fast and exact computation of the $\cL_2$-discrepancy without errors resulting from numerical quadrature methods. Thus, the $\cL_2$-discrepancy is an ideal candidate for the training objective of our machine learning approach.

\subsection{Extension to higher dimensions}\label{sec:higherdim}
In many practical problems, particularly in engineering and finance, the dimension $d$ of the problem can be very large. This necessitates extending low-discrepancy sequences to the high dimensional case of $d \gg 1$. However, it is known \cite{MORCAF1994, WANGSLOAN2008} that the $\cL_2$-discrepancy fails to identify superior distributional properties of low-discrepancy point sets over random samples as the dimension increases. Indeed, in high dimensions the classical $\cL_2$-discrepancy of low-discrepancy point sets behaves like $\mathcal{O}(1/\sqrt{N})$, the same as for random points, for moderate values of $N$, while an improved order close to $\mathcal{O}(1/N)$ can only be seen for extremely large $N$. Empirical evidence for these last claims can be found in the discrepancy plots contained in \cite{MORCAF1994}. 

To this end, we suggest to base our new training objective for higher-dimensional generation of low-discrepancy points on the Hickernell $\cL_p$-discrepancy \cite{hickernell1998generalized},
\begin{equation}
    D_{H,p}(\{\bX_i\}_{i=1}^N) = \left(\sum_{\emptyset \neq s \subseteq \{1,\dots,d\}} \cL^p_p(\{\bX^s_i\}_{i=1}^N)\right)^\frac{1}{p}, 
\end{equation}
where $\emptyset \neq s \subseteq \{1,\dots,d\}$ is a non-empty subset of coordinate indices, and  $\{\bX^s_i\}_{i=1}^N$ is the projection of $\{\bX_i\}_{i=1}^N$ onto $[0,1]^{|s|}$. Note that while we can again make use of Warnock's formula \eqref{eq:warnock} to compute $D_{H,2}$, it requires computing the sum of the $\cL_2$-discrepancy of $2^d-1$ projections, which already for $d=32$ is more than $1$B. This highlights the necessity of modifying $D_{H,2}$ in order for it to be used as a training objective in a machine learning framework. Therefore, we suggest to base the training objective on a modification of the Hickernell $\cL_p$-discrepancy via random projections,
\begin{equation}\label{eq:ansatz}
    \tilde{D}_{H,p,K}(\{\bX_i\}_{i=1}^N) = \left(\sum_{k=1}^K \cL^p_p(\{\bX_i^{s_k}\}_{i=1}^N)\right)^\frac{1}{p},
\end{equation}
where $\emptyset \neq s_k \sim \cP(\{1,\dots,d\})$ are randomly sampled subsets of coordinate indices for each $k=1,\dots,K$, thus requiring to compute the $\cL_p$-discrepancy only $K$ times. \rev{More specifically, we sample $s_k$ by first selecting the dimension $d_k$ of the projection uniformly at random, and then sampling the projection itself uniformly at random from the set of all $d_k$-dimensional projections. This method ensures that the dimensions of the projections are uniformly distributed during training. Alternatively, we could randomly sample the dimensions of the projections according to a binomial distribution.}

\paragraph{Generating problem-dependent point sets.}
As a further advantage to this 
framework, we highlight its inherent flexibility. Specifically, employing the modified Hickernell discrepancy as the training objective represents a \emph{first step towards an adaptive QMC sampling method tailored for specific problems}. It is widely recognized that for many problems, the effective dimension—essentially, the number of dimensions capturing the majority of the problem's variability—is often significantly lower than the nominal dimension; for full details, refer to \cite{CAF1997}. Therefore, during high-dimensional training, \rev{assuming that the important subsets of variables are known or identified in advance, e.g., by  functional ANOVA methods \cite{LEMIEUXOWEN2002}}, prioritizing sampling from specific lower dimensional projections will yield a $d$-dimensional point set that is highly uniformly distributed in those same projections identified during training. This approach effectively creates a custom-made point set, optimized for problems that primarily depend upon particular subsets of variables.

\section{Empirical results}
In this section, we present empirical results comparing MPMC points to current state-of-the-art low-discrepancy point sets. More concretely, we demonstrate superior distributional properties of MPMC over other low-discrepancy point sets mainly with respect to the star-discrepancy, $D^*$. 

\rev{As previously mentioned, computing the star-discrepancy is typically a challenging task. The \textit{DEM algorithm} \cite{DEM} denotes the fastest method to compute the exact star-discrepancy with a significantly reduced complexity, running in $\mathcal{O}(N^{1+d/2})$ time.} 
In all experiments, the results of which are presented shortly, to calculate the star discrepancy we either use a simple crude search or a parallelized version of the DEM algorithm from \cite{DEMparallel2023} to speed up calculation when necessary. 
The interested reader is recommended to consult \cite{DOERR2014} and references therein for more information on the calculation of the star-discrepancy, and indeed the computation of other discrepancy measures.

\subsection{Low-dimensional generation of MPMC points}
Here, we focus on generating MPMC points within a lower-dimensional setting particularly because this area has recently attracted significant attention \cite{CLEMENTDOERR2022, CLEMENTDOERR2024, CLEMENTOPTIMAL2023} providing a solid basis for comparison.

\begin{figure}[ht!]
\includegraphics[width=8.7cm]{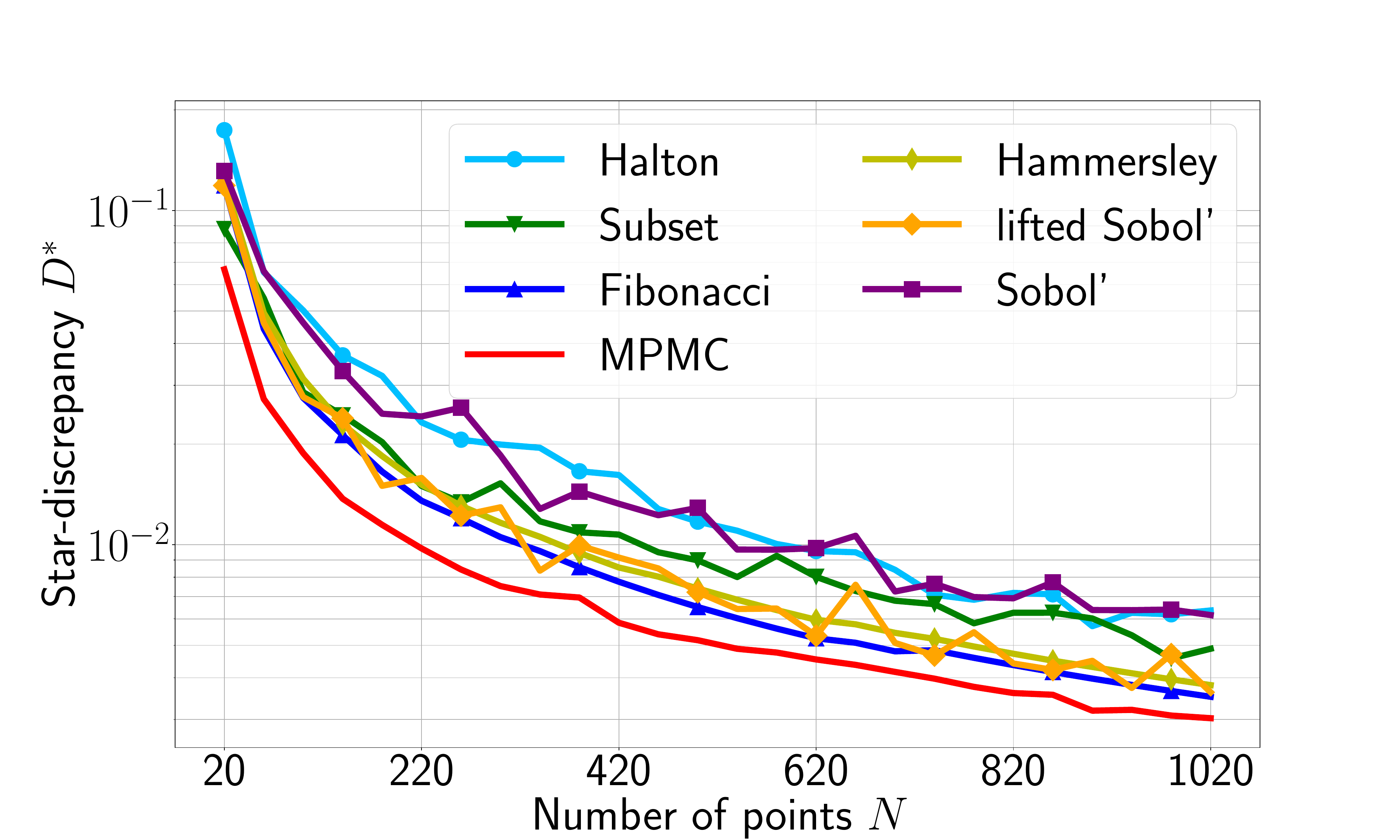}
\caption{Star-discrepancy $D^*$ of Halton, Sobol', \rev{lifted Sobol'}, Subset Selection, Hammersley, Fibonacci, and MPMC for increasing number of points $N=20,\dots,\rev{1020}$ in $d=2$.}
\label{fig:star_disc_plot}
\end{figure}

We compare the irregularity of our MPMC point sets with a truncation to the first $N$ points of the widely used Sobol' and Halton sequences. 
We note that MPMC points are sets optimized for $N$ chosen in advance, whereas a key advantage of the Sobol' and Halton sequences are that they are built for repeated sampling and retain a low-discrepancy for all values of $N$. Therefore, in addition, we provide comparison with state-of-the-art point sets derived from the subset selection method from \cite{CLEMENTDOERR2022, CLEMENTDOERR2024}, \rev{lifted Sobol'} \secrev{(the first $N$ terms of a one-dimensional Sobol' sequence concatenated with $i/N$ for $i \in \{0,\dots,N-1\}$)}, the Hammersley construction in base $1+\sqrt{2}$ as introduced in \cite{KIRKLEMIEUXWIART2023} and the Fibonacci set defined as $\left\{ (i/N, \{i\varphi\}) : i \in \{0, \ldots, N-1\} \right\}$ where $\varphi$ represents the golden ratio, the notation $\{ x \}$ denotes the fractional part of $x\in\mathbb{R}$. All four of these sets are recognized for having among the lowest star-discrepancy for given $N$ in two dimensions. \fref{fig:star_disc_plot} shows the star-discrepancy of MPMC, Sobol', Halton, subset selection, \rev{lifted Sobol'}, Hammersley and Fibonacci sets in two dimensions for increasing number of points $N=20,\dots,\rev{1020}$. We can see that MPMC significantly outperforms all other methods with respect to the star-discrepancy. In fact, the star-discrepancy of MPMC is on average \secrev{$1.3$} times smaller than that of the current state-of-the-art Fibonacci construction, and on average more than $2.2$ times smaller than Sobol' or Halton points. We provide the exact values of \fref{fig:star_disc_plot} in the Supporting Information (SI) such that it can serve as a benchmark for future methods. \rev{We further provide the $\cL_2$-discrepancy values of MPMC together with a discussion outlining their relevance in the SI.}

\subsection{Optimality of MPMC point sets}
Much of the past research on low-discrepancy point sets focused on achieving star-discrepancy with optimal asymptotic order in $N$ for implementation in quasi-Monte Carlo methods. However, there has been a recent surge in interest in finding point sets that minimize discrepancy for fixed $N$ and $d$. 
The main contribution in this direction was given in \cite{CLEMENTOPTIMAL2023}, where the authors constructed optimal star-discrepancy point sets in two and three dimensions. Naturally, we are interested in comparing MPMC points to these optimal formulations. 
The results of the optimal star-discrepancy comparison in two dimensions are presented in \Tref{table:2Dcomparison} and the three dimensional case is found in the SI.

\begin{table}[ht]
\centering
\caption{Comparison in two dimensions of MPMC star-discrepancy values against optimal sets and Fibonacci sets.}
\label{table:2Dcomparison}
\resizebox{\columnwidth}{!}{
\begin{tabular}{lccccccc}
\toprule
N & 1 & 2 & 3 & 4 & 5 & 6 & 7 \\
\midrule
Fibonacci & 1.0                  & 0.6909 & 0.5880 & 0.4910 & 0.3528 & 0.3183 & 0.2728 \\
Optimal   & \( 1/\varphi \) & 0.3660  & 0.2847 & 0.2500   & 0.2000  & 0.1667 & 0.1500 \\
\textbf{MPMC} & \( 1/\varphi \) & 0.3660 & 0.2847 & 0.2500 & 0.2000  & 0.1692 & 0.1508 \\
\midrule
\midrule
N & 8 & 9 & 10 & 11 & 12 & 13 & 14 \\
\midrule
Fibonacci  & 0.2553 & 0.2270 & 0.2042 & 0.1857 & 0.1702 & 0.1571 & 0.1459 \\
Optimal  & 0.1328 & 0.1235 & 0.1111 & 0.1030 & 0.0952 & 0.0889 & 0.0837 \\
\textbf{MPMC} & 0.1354 & 0.1240 & 0.1124 & 0.1058 & 0.0975 & 0.0908 & 0.0853  \\
\midrule
\midrule
N & 15 & 16 & 17 & 18 & 19 & 20 & \rev{21} \\
\midrule
Fibonacci & 0.1390 & 0.1486 & 0.1398 & 0.1320 & 0.1251 & 0.1188 & \rev{0.1131} \\
Optimal    & 0.0782 & 0.0739 & 0.0699 & 0.0666 & 0.0634 & 0.0604 & \rev{0.0580} \\
\textbf{MPMC} & 0.0794  & 0.0768 & 0.0731 & 0.0699 & 0.0668 & 0.0640 & \rev{0.0606} \\
\bottomrule
\end{tabular}
}
\end{table}
We can see that the star-discrepancy of MPMC points is very close to the star-discrepancy of the optimal point sets, and in fact match it exactly for small $N$. Moreover, the star-discrepancy of the Fibonacci set is far off the optimal values, i.e., approximately by a factor of $2$. Finally, it is worth highlighting, \rev{as reported in \cite{CLEMENTOPTIMAL2023}}, that the computation of the optimal points requires to solve a non-linear programming problem and takes approximately $18$ days to compute \secrev{for the case of $d=3$ and $N=8$.} \secrev{In contrast to that, MPMC was trained from scratch in $72$ seconds on an NVIDIA GeForce RTX 2080 Ti GPU for the same case.} 

\subsection{MPMC generation in high dimensions}

As discussed in Section \ref{sec:higherdim}, the efficacy of discrepancy measures to justly evaluate irregularity of distribution is flawed in higher dimensions. Therefore, to alternatively assess the quality of the distribution of higher dimensional point sets and sequences, motivated by the Koksma-Hlawka inequality \eqref{eq:KHinequality}, we will implement high dimensional MPMC points in an integral arising in a real-world problem from computational finance previously studied in \cite{LEMIEUXOWEN2002, WANGSLOAN2005, LEMIEUXFAURE2009}.

The primary goal is to accurately estimate the value at time $0$ of an Asian call option on an underlying asset that follows a log-normal distribution. Complete details of the problem formulation are provided in the SI. With our chosen parameters, this problem is known \cite{LEMIEUXOWEN2002} to exhibit more than 97\% of its variability in dimensions one, two, and three. 
Table \ref{table:comp_finance} shows the absolute errors observed when implementing \rev{Hammersley, lattice}, Sobol' and MPMC constructions. We train a 32-dimensional MPMC point set while emphasizing the $1$-$3$-dimensional projections as described in Section \ref{sec:higherdim}. \rev{Likewise, we utilize the custom QMC software LatNet Builder \cite{LatNet2020} to construct a rank-1 lattice by the component-by-component construction \cite{KUOJOE2002, DICK2015} placing importance on the $1$-$3$-dimensional projections.} We report the average absolute error of an MPMC training batch, which is selected based on the minimal Hickernell $\cL_2$-discrepancy restricted to $1$-$3$-dimensional projections.

\begin{table}[ht]
\centering
\caption{Approximation error of an Asian call option pricing of MPMC, \rev{ Hammersley, a rank-1 lattice}, and Sobol'.}
\label{table:comp_finance}
\begin{tabular}{lcccccc}
\toprule
N & 32 & 64 & 128 & 256 & 512 & \rev{1024}\\
\midrule
\rev{Hammersley} & \secrev{6.449} & \secrev{4.125} & \secrev{3.575} & \secrev{2.817} & \secrev{1.947} & \secrev{1.296}\\
\rev{Rank-1 lattice} & \secrev{5.636} & \secrev{4.638} & \secrev{1.331} & \secrev{2.151} & \secrev{0.180} & \secrev{0.203} \\
Sobol' & \secrev{1.235} & \secrev{1.373} & \secrev{0.965} & \secrev{0.623} & \secrev{0.497} & \secrev{0.290} \\
\textbf{MPMC} &  \secrev{1.402} & \secrev{0.831} & \secrev{0.512} & \secrev{0.250} & \secrev{0.120} & \secrev{0.055} \\
\bottomrule
\end{tabular}
\end{table}

\rev{The 32-dimensional MPMC point sets show significant enhancements over previous methods. This improvement is particularly notable at higher values of $N$, where MPMC outperforms rank-1 lattice by a factor of approximately $4$, Sobol' by a factor of $5$, and Hammersley by a factor of $24$. A particularly promising feature is the gain shown by MPMC points through targeted training, surpassing LatNet Builder's targeted rank-1 lattice.} The efficiency observed in this high-dimensional problem suggests a superior uniformity of MPMC points. In fact, the observed improvements may partially be explained by the uniformity held in lower-dimensional projections when compared to the traditional choices of QMC point sets. We refer to the SI for further discussion.

\subsection{Ablations}
Our proposed MPMC method is the result of several design choices, such as the type of input points, the deep learning model, and the training objective. In order to further justify our choices, we ablate several aspects of our MPMC framework illustrated by the following questions:

How does the graph structure influence the performance? To answer this, we compute the average $\cL_2$-discrepancy of several trained MPMC models for increasing values of the nearest neighbor radius $r$ in \eqref{eq:radius_graph} ranging from $0$ to $\sqrt{d}$ for different number of points $N=64,128,1024$, and plot the results in the SI. These results lead to two important observations. First, not using a graph structure at all, i.e., setting $r=0$ resulting in Deepsets \cite{zaheer2017deep}, significantly impairs the performance of MPMC, reaching average $\cL_2$-discrepancy values that are $9$ to $40$ times worse than using a graph structure. The second observation is that although the performance of MPMC is relatively stable for any choice of $r>0$, including the radius $r$ in hyperparameter tuning can help achieve point sets with minimal discrepancy.

What is the role of the GNN architecture used in MPMC? While we base MPMC on message-passing neural networks (MPNNs) \cite{mpnn}, other GNN architectures such as Graph Convolutional Networks (GCNs) \cite{gcn}, or Graph Attention Networks (GATs) can be used in this context as well. To check this, we train MPMC based on MPNNs, GCNs, and GATs for three different number of points $N=64,256,1024$ and show the $\cL_2$-discrepancy in the SI. Based on these results, we conclude that GCNs and GATs outperform each other based on the number of points $N$. At the same time, MPNNs consistently yield point sets with the lowest discrepancy values among all three considered GNN architectures. 

Does the choice of input point sets described in Section \ref{sec:inout_points} influence the performance of MPMC? To answer this, we train several MPMC models on all three different types of input points, i.e., random points, Sobol', and a randomized Sobol', where we choose $\xi \sim \mathcal{U}([0,0.1]^d)$ in \eqref{eq:randqmc}, for two different number of points $N=256,1024$. We report the average $\cL_2$-discrepancy of all trained MPMC models for increasing number of training steps in the SI. We observe that on average Sobol' and randomized Sobol' yield slightly lower discrepancy values and faster convergence compared to random points and thus lead to a more robust performance. However, we further note, that instead of averaging over all trained MPMC models, but instead choosing the single best model yield similar results for each input point type. 

\section{Discussion}
Low-discrepancy points play a central role in many applications in science and engineering. 
In this article, we have proposed MPMC, the first machine learning approach to generate new sets of low-discrepancy points. Inspired by the geometric nature of constructing such point sets, we base our MPMC approach on GNNs. Choosing an adequate training objective, i.e., closed-form solution of the $\cL_2$-discrepancy, we show that MPMC successfully transforms (random) input points into point sets with low discrepancy. Moreover, we extend this framework to higher dimensions, by training with an approximation of the Hickernell $\cL_2$-discrepancy. 
We further present an extensive empirical evaluation to illustrate different aspects of the proposed MPMC approach, highlighting the superior \rev{uniformity} properties of MPMC points compared to previous state-of-the-art methods. Finally, we carefully ablate key components of our MPMC model, yielding deeper empirical insights. 

MPMC represents a novel and efficient way of generating point sets with very low discrepancy. In fact, MPMC is \rev{empirically} shown to obtain optimal or near-optimal discrepancy \rev{for every dimension and the number of points for which the optimal discrepancy can be determined}. This is crucial for computationally expensive applications, where MPMC will lead to potentially significantly lower absolute errors compared to previous methods. Moreover, the generality of the MPMC framework allows for designing tailor-made QMC points that exploit specific structures of the problem at hand.

The aim of this paper was to generate point sets with low discrepancy for a fixed dimension and fixed number of points. On the other hand, many important applications require repeated sampling resulting in low-discrepancy sequences and not fixed point sets. Thus, one important aspect of future work will be to extend our MPMC point sets to MPMC sequences. 
\rev{Lastly, based on the superior discrepancy performance, we expect MPMC point sets to excel in various applications. Motivated by this, we would like to apply MPMC to various problems in science and engineering as future work.}

\clearpage
\acknow{The authors would like to thank François Clément (Sorbonne Université, CNRS) for several helpful discussions, and for providing computer code for the further visual and empirical insights contained in the SI. 
This research was supported in part by the AI2050 program at Schmidt Futures (grant G-22-63172), the Boeing Company, and the United States Air Force Research Laboratory and the United States Air Force Artificial Intelligence Accelerator and was accomplished under cooperative agreement number FA8750-19-2-1000. 
The work of TKR is supported by Postdoc.Mobility grant P500PT-217915 from the Swiss National Science Foundation. The work of NK and CL is supported by the Natural Science and Engineering Research Council of Canada (NSERC) via grant 238959. MB is supported in part by EPSRC Turing AI World-Leading Research Fellowship No. EP/X040062/1.
}

\showacknow{} 



\bibliography{refs}
\appendix
\renewcommand{\thesection}{\Alph{section}}
\renewcommand{\thesubsection}{\thesection.\arabic{subsection}}

\onecolumn
\begin{center}
{\bf Supporting Information (SI) for}\\
Message-Passing Monte Carlo: Generating low-discrepancy point sets via Graph Neural Networks
\end{center}

\section{Training details}
All experiments have been run on NVIDIA GeForce RTX 2080 Ti, GeForce RTX 3090, TITAN RTX and Quadro RTX 6000 GPUs. Each model was trained for initial $100$k training steps, after which the learning rate was reduced by a factor of $10$ whenever the discrepancy measure of the output point sets did not improve for a total of $2$k training steps evaluated after every $100$ training steps. The training was stopped once the learning rate reached a value less than $10^{-6}$. Moreover, the hyperparameters of the model were tuned based on random search according to \Tref{tab:hp_ranges}, which shows the search-space of each hyperparameter as well as the random distribution used to sample from it.

\begin{table*}[h!]
    \centering
    \caption{Hyperparameter search-space and random distributions to sample from it.}
    \label{tab:hp_ranges}
    \begin{tabular}{lcc}
    \toprule 
         &
         \textbf{range} &
         \textbf{distribution} \\
         \midrule
         
         learning rate &
         $[10^{-4},10^{-2}]$ &
         log uniform
         \\
         
         hidden size $m_0=m_1=\dots=m_L$ &
         $\{32, 64, 128, 256\}$ &
         disc. uniform \\

         number of GNN layers $L$ &
         $\{1,2,\dots,10\}$ &
         disc. uniform \\

         size of mini-batches &
         $\{8,16,32\}$ &
         disc. uniform \\
         
         weight decay &
         $[10^{-8},10^{-2}]$ &
         log uniform
         \\
         
        \bottomrule
    \end{tabular}
\end{table*}

\section{On the Asian Option Pricing Experiment}\label{sec:finance}

We describe the problem of estimating the value at time $0$ of an Asian call option on an underlying asset in detail. The results of which are presented in the main text as Table 2.

The main goal is to estimate an expectation of the form, $$C_0 = \mathbb{E} \left[ e^{-rT} \left( \frac{1}{d} \sum_{j=1}^{d} S(u_j) - K \right)^+ \right].$$ We let \( T \) be the expiration time of the contract, \( K \) the strike price, for $Z \sim N(0,1)$ let $S(u) = S(0)e^{\left(r-\frac{\sigma^2}{2}\right)u+\sigma\sqrt{u}Z}\) be the price of the underlying asset at time \( u \), and \( 0 < u_1 < \ldots < u_d = T \) are \( d \) times at which the asset price is observed in order to compute the average used in the option pricing formula. The expectation is taken under the risk-neutral probability measure. Finally, \( r \) is the risk-free rate, the notation \( x^+ \) means \( \max(0, x) \), $\Phi^{-1}$ is the inverse CDF of the standard normal distribution and $\Delta_l = u_l - u_{l-1}$. Assuming the stock price follows a geometric Brownian motion with volatility $\sigma$, it can be shown that this expectation can be written as follows:
\begin{equation*}
C_0 = e^{-rT} \int_{[0,1]^d} \left( \frac{1}{d} \sum_{j=1}^{d} S(0)e^{(r-\frac{\sigma^2}{2})u_j+\sigma  \sum_{l=1}^{j} \sqrt{\Delta_l} \Phi^{-1}(x_l)} - K \right)^+ \, dx_1 \ldots dx_d.
\end{equation*}
In our simulations, the true value \secrev{$C_0 = 7.06574$} was calculated in advance via QMC simulation with $2$M Sobol' points with the following set of parameters: $S(0) = 50, T = 1 \text{ year}, r = 0.05, \sigma = 0.3, K=45$ and $d=32$.

\secrev{Further, for each $N$, the generating vector for the rank-1 lattice is produced from LatNet builder from the command line by the following syntax: 
\noindent
\texttt{latnetbuilder -t lattice -c ordinary -s N -d 32 -e full-CBC -f CU:P2 -q 2 --weights file:/path/to/weights.txt.}
\noindent
The file weights.txt contains order-dependent weights of $10$ on projection orders 1, 2 and 3 and otherwise a default weight of $0.001$.}

\section{\rev{On the $\mathcal{L}_2$-discrepancy}}\label{sec:optimalL2}
\begin{figure}[ht!]
\centering
\includegraphics[width=0.6\textwidth]{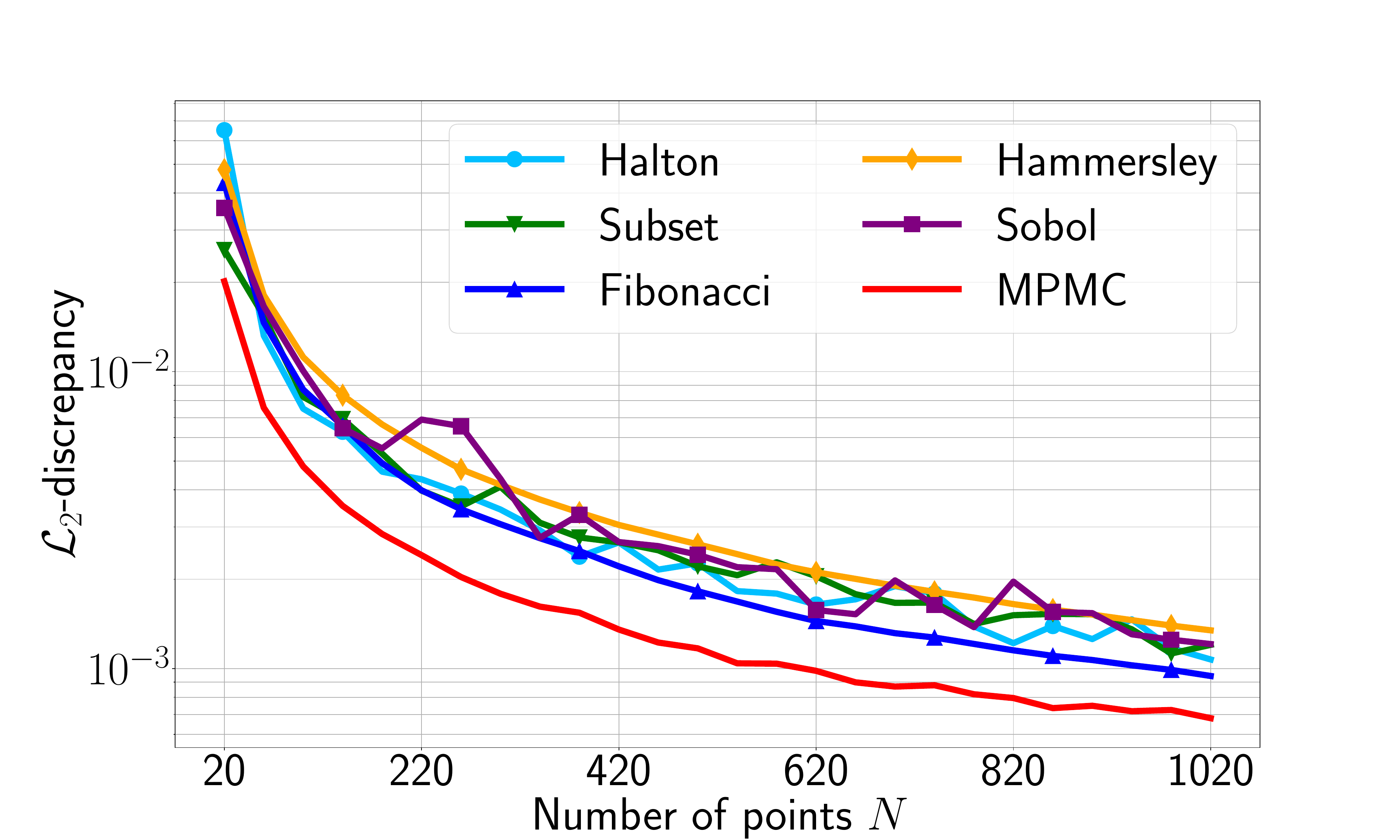}
\caption{\rev{$\mathcal{L}_2$-discrepancy of Halton, Sobol', Subset Selection, Hammersley, Fibonacci, and MPMC for increasing number of points $N=20,\dots,1020$ in $d=2$.}}
\label{fig:L2_disc_plot}
\end{figure}

\rev{The $\cL_2$-discrepancy is a well-researched and widely used measure of distribution irregularity, consistently attracting attention from the QMC research community \cite{BORDA2024, KIRKPAUSINGER2023, KRITZ2022}. In our MPNN framework, we utilize the $\mathcal{L}_2$-discrepancy function as a training objective because of its efficient computation and differentiable formulation to be used in the gradient-based learning. This allows the generation of point sets with very small star discrepancy. As a by-product, we also create point sets with small $\mathcal{L}_2$-discrepancy. \fref{fig:L2_disc_plot} shows the $\cL_2$-discrepancy values for increasing number of points $N=20,\dots,1020$ for MPMC, Halton, Sobol', Subset Selection method, Hammersley, and Fibonacci in $2$ dimensions. We further present the numerical values of \fref{fig:L2_disc_plot} in \Tref{tab:L2_disc}. We can see that MPMC consistently obtains the lowest $\cL_2$-discrepancy values for all number of points $N$. Moreover, MPMC significantly outperforms the previous state-of-the-art method based on Fibonacci point sets. We highlight that the performance gap between MPMC and any other method considered here is even wider for the $\cL_2$-discrepancy than for the star-discrepancy $D^*$ in Figure 4 of the main text. This can be explained by the fact that MPMC is trained to minimize the $\cL_2$-discrepancy directly, and thus by design optimizes the $\cL_2$-discrepancy instead of the star-discrepancy $D^*$.}

\begin{table}[ht!]
\centering
\caption{\rev{$\mathcal{L}_2$-discrepancy values for Halton, Sobol', Subset Selection, Hammersley, Fibonacci, and MPMC for different number of points $N=20,\dots,1020$ in $d=2$.}}
\label{tab:L2_disc}
\resizebox{\textwidth}{!}{
\begin{tabular}{lcccccccccccccc}
\toprule
\rev{N} & \rev{20 }& \rev{60 }& \rev{100} & \rev{140 }& \rev{180} & \rev{220 }& \rev{260 }& \rev{300} & \rev{340 }& \rev{380} & \rev{420} & \rev{460} & \rev{500} \\
\midrule
\rev{Halton} &  
\rev{ 0.06511 } &
\rev{ 0.01323 } &
\rev{ 0.00751 } &
\rev{ 0.00627 } &
\rev{ 0.00460 } &
\rev{ 0.00434 } &
\rev{ 0.00389 } &
\rev{ 0.00343 } &
\rev{ 0.00292 } &
\rev{ 0.00238 } &
\rev{ 0.00266 } &
\rev{ 0.00215 } &
\rev{ 0.00226 } \\
\rev{Sobol'} &  
\rev{ 0.03564 } &
\rev{ 0.01660 } &
\rev{ 0.01006 } &
\rev{ 0.00645 } &
\rev{ 0.00551 } &
\rev{ 0.00690 } &
\rev{ 0.00655 } &
\rev{ 0.00435 } &
\rev{ 0.00276 } &
\rev{ 0.00330 } &
\rev{ 0.00267 } &
\rev{ 0.00258 } &
\rev{ 0.00242 } \\
\rev{Subset Selection} & 
\rev{ 0.02569 } &
\rev{ 0.01541 } &
\rev{ 0.00823 } &
\rev{ 0.00693 } &
\rev{ 0.00530 } &
\rev{ 0.00397 } &
\rev{ 0.00352 } &
\rev{ 0.00410 } &
\rev{ 0.00310 } &
\rev{ 0.00276 } &
\rev{ 0.00266 } &
\rev{ 0.00250 } &
\rev{ 0.00221 } \\
\rev{Hammersley} & \rev{ 0.04796 } &
\rev{ 0.01812 } &
\rev{ 0.01119 } &
\rev{ 0.00833 } &
\rev{ 0.00664 } &
\rev{ 0.00554 } &
\rev{ 0.00469 } &
\rev{ 0.00416 } &
\rev{ 0.00371 } &
\rev{ 0.00335 } &
\rev{ 0.00305 } &
\rev{ 0.00283 } &
\rev{ 0.00262 } \\
\rev{Fibonacci} &  
\rev{ 0.04324 } &
\rev{ 0.01465 } &
\rev{ 0.00870 } &
\rev{ 0.00657 } &
\rev{ 0.00492 } &
\rev{ 0.00399 } &
\rev{ 0.00344 } &
\rev{ 0.00306 } &
\rev{ 0.00275 } &
\rev{ 0.00249 } &
\rev{ 0.00221 } &
\rev{ 0.00198 } &
\rev{ 0.00182 } \\
\rev{\textbf{MPMC}} & 
\rev{ 0.02016 } &
\rev{ 0.00756 } &
\rev{ 0.00479 } &
\rev{ 0.00353 } &
\rev{ 0.00284 } &
\rev{ 0.00241 } &
\rev{ 0.00203 } &
\rev{ 0.00179 } &
\rev{ 0.00162 } &
\rev{ 0.00154 } &
\rev{ 0.00135 } &
\rev{ 0.00122 } &
\rev{ 0.00117 } \\
\midrule
\midrule
\rev{N }& \rev{540 }&\rev{ 580} &\rev{ 620 }&\rev{ 660} & \rev{700} &\rev{ 740 }& \rev{780} &\rev{ 820 }& \rev{860} &\rev{ 900 }& \rev{940} & \rev{980} & \rev{1020} \\
\midrule
\rev{Halton} & 
\rev{ 0.00182 } &
\rev{ 0.00179 } &
\rev{ 0.00164 } &
\rev{ 0.00171 } &
\rev{ 0.00190 } &
\rev{ 0.00179 } &
\rev{ 0.00139 } &
\rev{ 0.00122 } &
\rev{ 0.00139 } &
\rev{ 0.00126 } &
\rev{ 0.00147 } &
\rev{ 0.00117 } &
\rev{ 0.00107 } &\\
\rev{Sobol'} & 
\rev{ 0.00220 } &
\rev{ 0.00216 } &
\rev{ 0.00157 } &
\rev{ 0.00152 } &
\rev{ 0.00198 } &
\rev{ 0.00164 } &
\rev{ 0.00138 } &
\rev{ 0.00196 } &
\rev{ 0.00155 } &
\rev{ 0.00154 } &
\rev{ 0.00130 } &
\rev{ 0.00125 } &
\rev{ 0.00121 } \\
\rev{Subset Selection} & 
\rev{ 0.00206 } &
\rev{ 0.00228 } &
\rev{ 0.00205 } &
\rev{ 0.00178 } &
\rev{ 0.00166 } &
\rev{ 0.00167 } &
\rev{ 0.00141 } &
\rev{ 0.00151 } &
\rev{ 0.00153 } &
\rev{ 0.00152 } &
\rev{ 0.00136 } &
\rev{ 0.00112 } &
\rev{ 0.00120 } \\
\rev{Hammersley} & 
\rev{ 0.00243 } &
\rev{ 0.00224 } &
\rev{ 0.00211 } &
\rev{ 0.00200 } &
\rev{ 0.00190 } &
\rev{ 0.00181 } &
\rev{ 0.00173 } &
\rev{ 0.00165 } &
\rev{ 0.00158 } &
\rev{ 0.00152 } &
\rev{ 0.00146 } &
\rev{ 0.00140 } &
\rev{ 0.00135 } \\
\rev{Fibonacci} & 
\rev{ 0.00168 } &
\rev{ 0.00155 } &
\rev{ 0.00145 } &
\rev{ 0.00139 } &
\rev{ 0.00132 } &
\rev{ 0.00127 } &
\rev{ 0.00121 } &
\rev{ 0.00115 } &
\rev{ 0.00110 } &
\rev{ 0.00107 } &
\rev{ 0.00103 } &
\rev{ 0.00099 } &
\rev{ 0.00094 } \\
\rev{\textbf{MPMC}} &
\rev{ 0.00104 } &
\rev{ 0.00104 } &
\rev{ 0.00098 } &
\rev{ 0.00090 } &
\rev{ 0.00087 } &
\rev{ 0.00088 } &
\rev{ 0.00082 } &
\rev{ 0.00080 } &
\rev{ 0.00074 } &
\rev{ 0.00075 } &
\rev{ 0.00072 } &
\rev{ 0.00072 } &
\rev{ 0.00068 } \\

\bottomrule
\end{tabular}
}
\end{table}

\section{Further empirical insights}

\subsection{\secrev{Post-construction enhancements}}
\secrev{Classical quasi-Monte Carlo (QMC) constructions have known limitations, including poor uniformity in low-dimensional projections of high-dimensional point sets (see Figure \ref{fig:projections5and6} and \ref{fig:projections26and27}). To mitigate these issues, significant amount of research has concentrated on incorporating randomness into these deterministic constructions, as discussed in \cite{OKTEN2012, HICKWANG2000, ECUYER2002}. This approach also offers the advantage of using repeated uniform sampling -- known as randomized quasi-Monte Carlo (RQMC) -- for straightforward unbiased error estimation.
}

\secrev{One successful method is Owen scrambling in base $b$ \cite{OWEN1995} which involves sequentially applying uniformly chosen permutations to the base $b$ digits of each point in the set. This method has proven to be very popular and highly effective. Another simpler technique is the random shift modulo $1$ \cite{PATTERSON1976}, where a vector in $[0,1]^d$ is chosen uniformly at random and added (modulo 1) to each point in the set.}

\secrev{In this section, we test how these enhancements improve the results of selected QMC constructions in the Asian option pricing application presented in Section 3.C. of the main text (i.e., Table 2 of the main text). To this end, we start by applying scrambling (in the natural base $2$) to the Sobol' sequence and present the mean absolute error (MAE) in \Tref{tab:MAE_randomization}. We can see that on average scrambling massively improves the absolute error of Sobol', i.e., by a factor of more than $10$. While Owen scrambling in base $b$ can be applied to an arbitrary point set, the choice of $b$ is paramount to ensure that the low-discrepancy property is preserved after randomization. Since the correct choice of the base is not immediately apparent, scrambling is not directly applicable to MPMC and a direct comparison involving both methods leveraging the scrambling enhancement is not possible. However, uniform random shifting (modulo 1) can be applied to any QMC construction, including MPMC. Therefore, we present the results of randomly shifted MPMC in \Tref{tab:MAE_randomization}. We can see that randomly shifting on average leads to a lower absolute error compared to MPMC without random shifting for smaller number of points $N$. For large $N=1024$, however, MPMC with random shifting appears to perform worse than MPMC without random shifting. \emph{This highlights the importance of developing suitable randomization techniques specifically tailored for MPMC, a topic we plan to focus on in future research.}}
\begin{table}[h!]
\centering
\caption{\secrev{Errors of Sobol', scrambled Sobol', MPMC and randomly shifted MPMC for the Asian option pricing experiment in Section 3.C. of the main text. MPMC and Sobol' errors are provided as absolute errors and taken from Table 2 in the main text, while shifted MPMC and scrambled Sobol' are provided as mean absolute errors (MAE).}}
\label{tab:MAE_randomization}
\begin{tabular}{lcccccc}
\toprule
\secrev{N} & \secrev{32} & \secrev{64} & \secrev{128} & \secrev{256} & \secrev{512} & \secrev{1024} \\
\midrule
\secrev{Sobol'} & \secrev{1.235} & \secrev{1.373} & \secrev{0.965} & \secrev{0.623} & \secrev{0.497} & \secrev{0.290} \\
\secrev{Scrambled Sobol'} & \secrev{0.516} &  \secrev{0.169} &  \secrev{0.076} &  \secrev{0.064} &  \secrev{0.040} &  \secrev{0.022} \\
\secrev{MPMC} & \secrev{1.402} & \secrev{0.831} & \secrev{0.512} & \secrev{0.250} & \secrev{0.120} & \secrev{0.055} \\
\secrev{Shifted MPMC} & \secrev{0.521} & \secrev{0.310} & \secrev{0.188} & \secrev{0.128} & \secrev{0.082} & \secrev{0.061} \\

\bottomrule
\end{tabular}
\end{table}

\begin{figure}[h!]
\begin{minipage}{\textwidth}
\begin{minipage}{.33\textwidth}
\includegraphics[width=1.\textwidth]{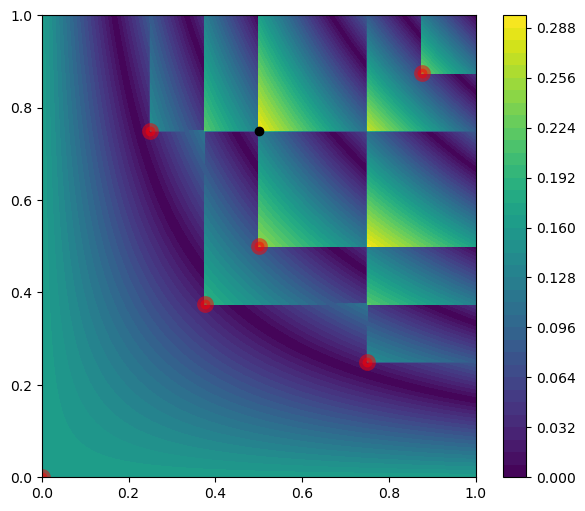}
\end{minipage}%
\hfill
\begin{minipage}{.33\textwidth}
\includegraphics[width=1.\textwidth]{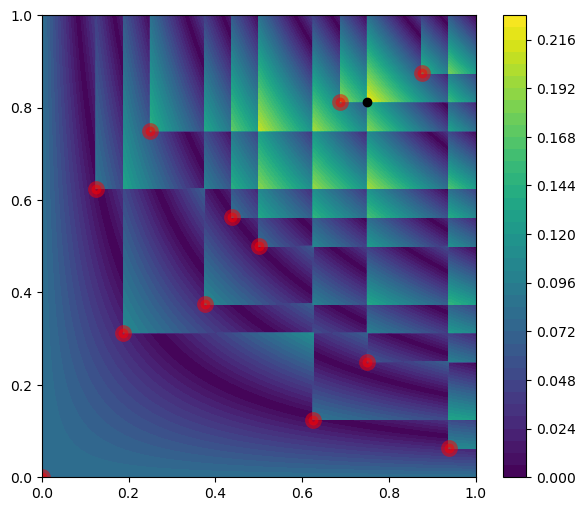}
\end{minipage}%
\hfill
\begin{minipage}{.33\textwidth}
\includegraphics[width=1.\textwidth]{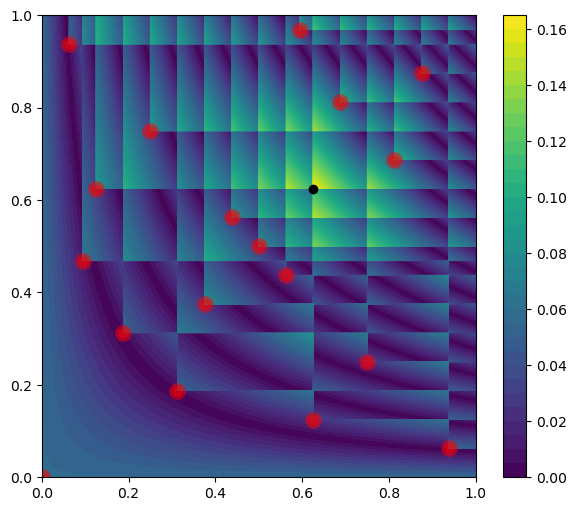}
\end{minipage}
\end{minipage}

\begin{minipage}{\textwidth}
\begin{minipage}{.33\textwidth}
\includegraphics[width=1.\textwidth]{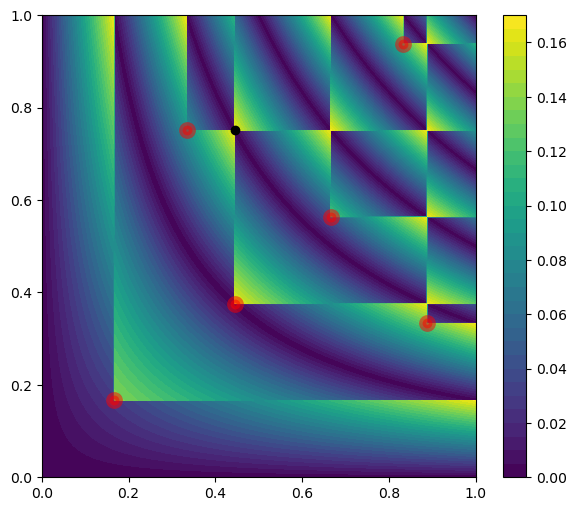}
\end{minipage}%
\hfill
\begin{minipage}{.33\textwidth}
\includegraphics[width=1.\textwidth]{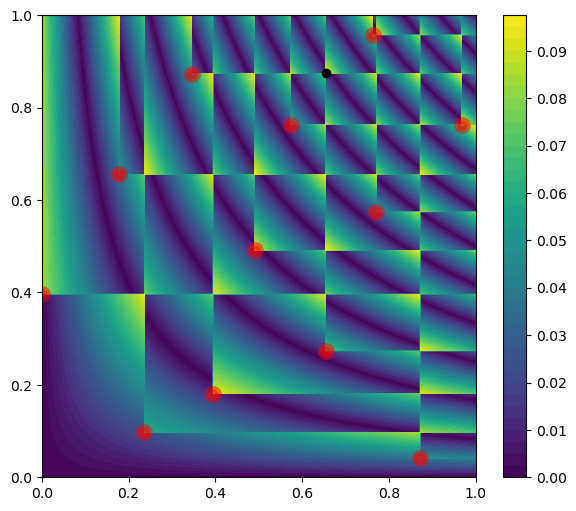}
\end{minipage}%
\hfill
\begin{minipage}{.33\textwidth}
\includegraphics[width=1.\textwidth]{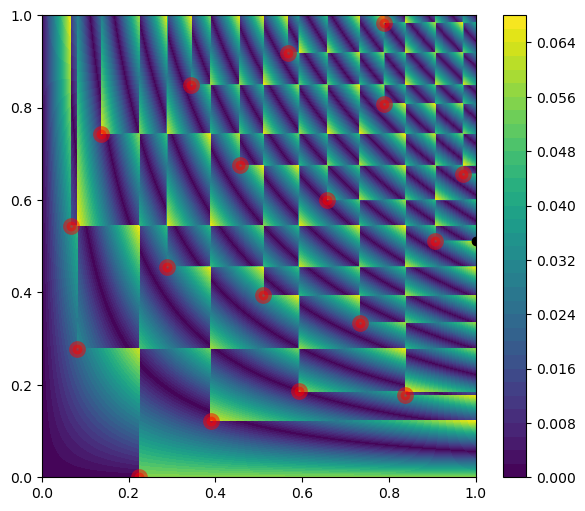}
\end{minipage}
\end{minipage}

\begin{minipage}{\textwidth}
\begin{minipage}{.33\textwidth}
\includegraphics[width=1.\textwidth]{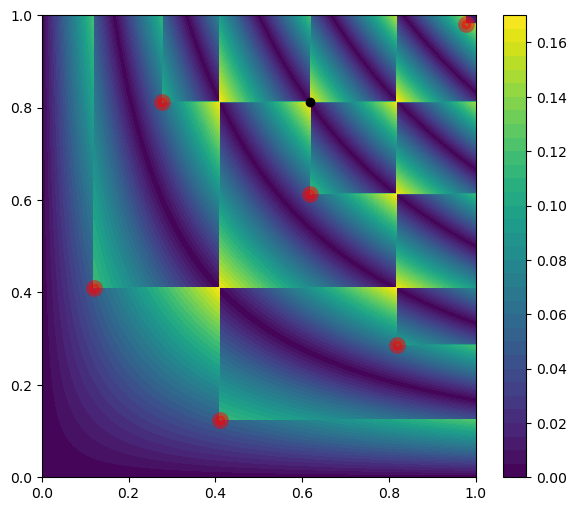}
\end{minipage}%
\hfill
\begin{minipage}{.33\textwidth}
\includegraphics[width=1.\textwidth]{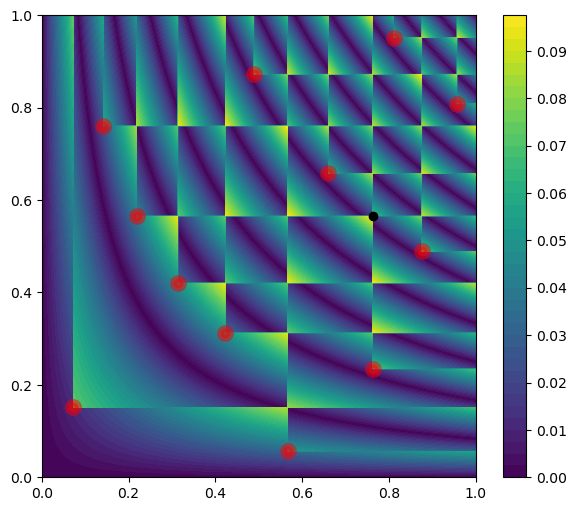}
\end{minipage}%
\hfill
\begin{minipage}{.33\textwidth}
\includegraphics[width=1.\textwidth]{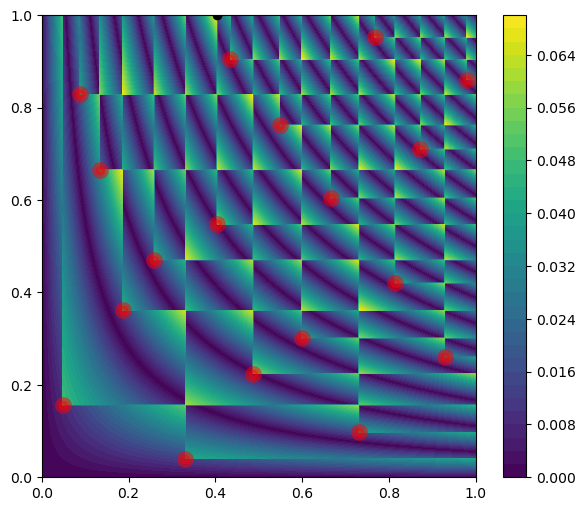}
\end{minipage}
\end{minipage}

\caption{Local discrepancy plots for Sobol' sequence (\textbf{top}), optimal point sets obtained in \cite{CLEMENTOPTIMAL2023} (\textbf{middle row}), and MPMC point sets (\textbf{bottom}) for $N=6$ points ({left}), $N=12$ points ({middle column}), and $N=18$ points ({right}).}
\label{fig:dim2_Sobol_localdisc}
\end{figure}

\subsection{Structure of MPMC Points}
Initially explored in \cite{CLEMENTOPTIMAL2023}, the authors provide insights into the configurations of two dimensional point sets that achieve optimal star-discrepancy by providing visualizations of the local discrepancy within the unit square. Providing equivalent comparisons, Figure \ref{fig:dim2_Sobol_localdisc} displays the local discrepancy plots for Sobol' sequences, optimal point sets and MPMC points. Each plot has its own color scale where darker areas indicate lower local discrepancy values, and brighter areas denote higher values. The presence of a black dot in each plot marks the point of maximum local discrepancy, i.e., the explicit anchored test box where the star-discrepancy is obtained. Additionally, regions of high local discrepancy form bright triangular regions whose corner is either directed toward the upper right corner to represent open boxes with too few points, or angled toward the lower left, indicating closed boxes with an excess of points. For instance, the Sobol' plots exclusively show closed overfilled boxes, with bright triangles pointing downward and leftward toward the origin. 

A visual comparison reveals structural similarities between the optimal sets and the MPMC points, suggesting a more balanced distribution of local discrepancy values across the unit square, with both open and closed boxes appearing in the plots. Interestingly, this similar structure emerges despite the sets consisting of quite different exact point values.

In conclusion, it seems evident that the GNN captures an essential underlying local discrepancy structure, which is key for minimizing star-discrepancy.

\subsection{Projections of MPMC points}

\begin{figure}[h!]
    \centering
    \includegraphics[width=\textwidth]{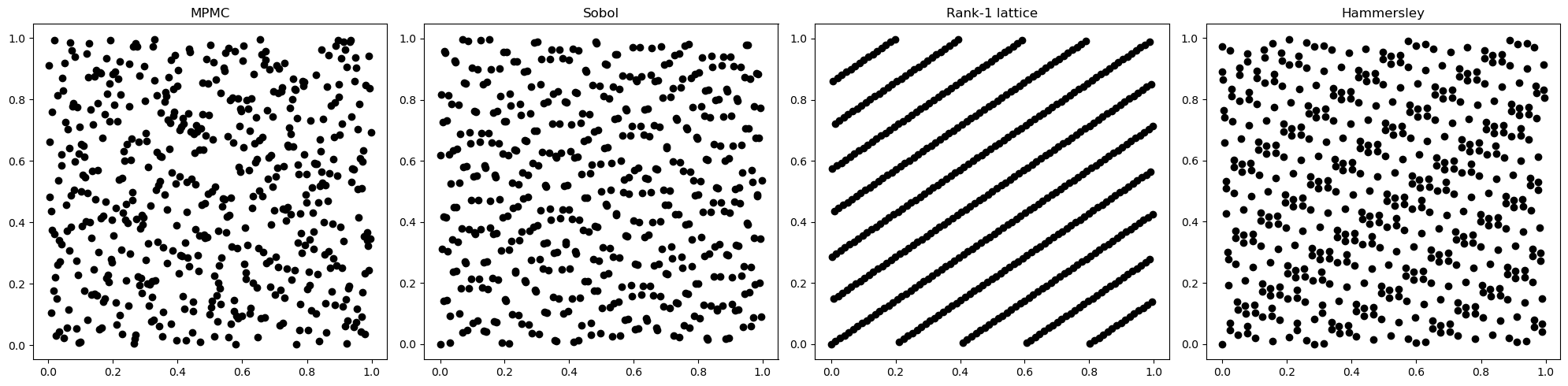}
    \caption{Projections of the $5$-th and $6$-th coordinates of 32-dimensional \rev{MPMC, Sobol', rank-1 lattice and Hammersley} with $N=512$ (left to right).}
    \label{fig:projections5and6}
\end{figure}

\begin{figure}[h!]
    \centering
    \includegraphics[width=\textwidth]{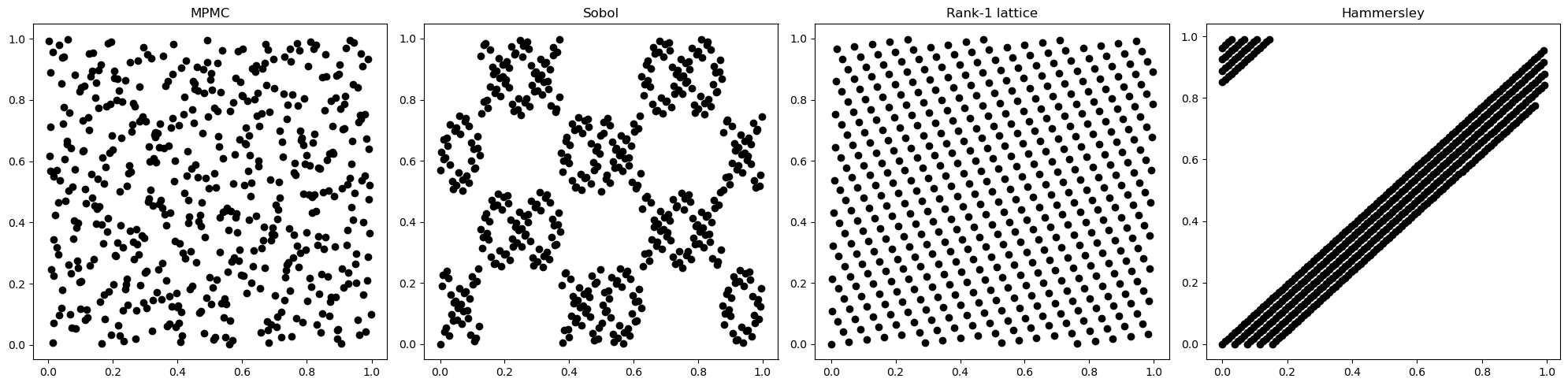}
    \caption{Projections of the $26$-th and $27$-th coordinates of 32-dimensional \rev{MPMC, Sobol', rank-1 lattice and Hammersley} with $N=512$ (left to right).}
    \label{fig:projections26and27}
\end{figure}

As noted in the main text, when applied to the computational finance integral described in Section \ref{sec:finance} to estimate the value of an Asian call option, the MPMC points far outperform the Sobol' or Hammersley in terms of approximation accuracy. A significantly important factor for the success of QMC methods in high dimensional application is the quality of the distribution in the lower dimensional projections of the QMC point set. See \cite{LEMIEUXOWEN2002} for a more comprehensive discussion. Figure \ref{fig:projections5and6} and \ref{fig:projections26and27} display the $5$-th and $6$-th, and $26$-th and $27$-th coordinate projections respectively of the MPMC points, Sobol'\rev{, rank-1 lattice and Hammersley} constructed in $32$ dimensions. Visual inspection reveals that the projections seem to be just as uniformly distributed in the $5$-th and $6$-th dimensions and notably more evenly distributed as the dimension increases to $26$ and $27$.  At these higher dimensions, we start noticing some undesired correlations in the coordinates of the Sobol' and \rev{Hammersley constructions}, however, fortunately the MPMC construction does not exhibit this problematic feature displaying no significant correlation, clustering, or sparsity; the MPMC point sets appear random yet maintain a high degree of uniformity. This characteristic is particularly advantageous for tackling high-dimensional problems.

\subsection{Optimality of MPMC in Three Dimensions}
\Tref{table:3Dcomparison} shows the star-discrepancy of MPMC in three dimensions for $N=1,2,\dots,8$ number of points, as well as the optimal star-discrepancy values obtained from \cite{CLEMENTOPTIMAL2023}. We can see that MPMC obtains again near-optimal star-discrepancy for all choices of $N$.
\begin{table}[h!]
\centering
\caption{Comparison in three dimensions of MPMC points star-discrepancy values against optimal sets.}
\label{table:3Dcomparison}
\begin{tabular}{lcccccccc}
\toprule
N & 1 & 2 & 3 & 4 & 5 & 6 & 7 & 8 \\
\midrule
Optimal  & 0.6823 & 0.4239 & 0.3445 & 0.3038 & 0.2618 & 0.2326 & 0.2090 & 0.1875 \\
\textbf{MPMC}  & 0.6833 & 0.4239 & 0.3491 & 0.3071 & 0.2669 & 0.2371 & 0.2158 & 0.1993 \\
\bottomrule
\end{tabular}
\end{table}

\subsection{Star-discrepancy values of Figure 4 in the main text}
From the main text, we present the exact numerical values of the star-discrepancy for Halton, Sobol', Subset Selection, Hammersley, \rev{lifted Sobol'}, Fibonacci, and MPMC points as a benchmark for future methods. 

\begin{table}[h!]
\centering
\caption{Star-discrepancy values of Figure 4 in the main text for Halton, Sobol', Subset Selection, Hammersley, \rev{lifted Sobol'}, Fibonacci, and MPMC.}
\label{table:3Dcomparison}
\resizebox{\textwidth}{!}{
\begin{tabular}{lcccccccccccccc}
\toprule
N & 20 & 60 & 100 & 140 & 180 & 220 & 260 & 300 & 340 & 380 & 420 & 460 & 500 \\
\midrule
Halton &  
0.17384  &
0.06535  &
0.05024  &
0.03686  &
0.03200  &
0.02323  &
0.02062  &
0.01994  &
0.01950  &
0.01659  &
0.01617  &
0.01279  &
0.01172  \\
Sobol' &  
0.13125  &
0.06583  &
0.04617  &
0.03306  &
0.02466  &
0.02420  &
0.02571  &
0.01851  &
0.01280  &
0.01442  &
0.01326  &
0.01225  &
0.01289  \\
Subset Selection &  
0.08799  &
0.05469  &
0.02860  &
0.02439  &
0.02028  &
0.01499  &
0.01339  &
0.01527  &
0.01175  &
0.01089  &
0.01073  &
0.00950  &
0.00898  \\
Hammersley & 
0.12304  &
0.04941  &
0.03136  &
0.02292  &
0.01842  &
0.01512  &
0.01308  &
0.01164  &
0.01056  &
0.00945  &
0.00855  &
0.00803  &
0.00739  \\
\rev{Lifted Sobol'} & 
\rev{ 0.11875 } &
\rev{ 0.04609 } &
\rev{ 0.02766 } &
\rev{ 0.02388 } &
\rev{ 0.01501 } &
\rev{ 0.01584 } &
\rev{ 0.01221 } &
\rev{ 0.01294 } &
\rev{ 0.00837 } &
\rev{ 0.00994 } &
\rev{ 0.00915 } &
\rev{ 0.00849 } &
\rev{ 0.00721 } \\
Fibonacci & 
0.11885  &
0.04422  &
0.02749  &
0.02128  &
0.01655  &
0.01354  &
0.01200  &
0.01054  &
0.00957  &
0.00857  &
0.00775  &
0.00708  &
0.00651  \\
\textbf{MPMC} & 
0.06664  &
0.02729  &
0.01879  &
0.01373  &
0.01147  &
0.00975  &
0.00843  &
0.00752  &
0.00710  &
0.00695  &
0.00584  &
0.00540  &
0.00518  \\
\midrule
\midrule
N & \rev{540} & \rev{580} & \rev{620} & \rev{660} & \rev{700} & \rev{740} & \rev{780} & \rev{820} & \rev{860} & \rev{900} & \rev{940} & \rev{980} & \rev{1020} \\
\midrule
Halton &  \rev{ 0.01101 } &
\rev{ 0.01005 } &
\rev{ 0.00957 } &
\rev{ 0.00949 } &
\rev{ 0.00841 } &
\rev{ 0.00709 } &
\rev{ 0.00685 } &
\rev{ 0.00718 } &
\rev{ 0.00710 } &
\rev{ 0.00571 } &
\rev{ 0.00626 } &
\rev{ 0.00619 } &
\rev{ 0.00636 }
\\
Sobol' &  
\rev{ 0.00967 } &
\rev{ 0.00967 } &
\rev{ 0.00977 } &
\rev{ 0.01064 } &
\rev{ 0.00724 } &
\rev{ 0.00765 } &
\rev{ 0.00697 } &
\rev{ 0.00691 } &
\rev{ 0.00772 } &
\rev{ 0.00638 } &
\rev{ 0.00637 } &
\rev{ 0.00640 } &
\rev{ 0.00616 } \\
Subset Selection &  
\rev{ 0.00800 } &
\rev{ 0.00927 } &
\rev{ 0.00802 } &
\rev{ 0.00727 } &
\rev{ 0.00680 } &
\rev{ 0.00664 } &
\rev{ 0.00582 } &
\rev{ 0.00626 } &
\rev{ 0.00626 } &
\rev{ 0.00602 } &
\rev{ 0.00536 } &
\rev{ 0.00457 } &
\rev{ 0.00488 } \\
Hammersley &  
\rev{ 0.00685 } &
\rev{ 0.00637 } &
\rev{ 0.00596 } &
\rev{ 0.00578 } &
\rev{ 0.00545 } &
\rev{ 0.00523 } &
\rev{ 0.00496 } &
\rev{ 0.00472 } &
\rev{ 0.00450 } &
\rev{ 0.00431 } &
\rev{ 0.00413 } &
\rev{ 0.00396 } &
\rev{ 0.00380 }\\
\rev{Lifted Sobol'} & 
\rev{ 0.00642 } &
\rev{ 0.00645 } &
\rev{ 0.00534 } &
\rev{ 0.00759 } &
\rev{ 0.00509 } &
\rev{ 0.00466 } &
\rev{ 0.00547 } &
\rev{ 0.00441 } &
\rev{ 0.00422 } &
\rev{ 0.00450 } &
\rev{ 0.00373 } &
\rev{ 0.00471 } &
\rev{ 0.00362 } \\
Fibonacci & 
\rev{ 0.00603 } &
\rev{ 0.00561 } &
\rev{ 0.00525 } &
\rev{ 0.00509 } &
\rev{ 0.00480 } &
\rev{ 0.00484 } &
\rev{ 0.00459 } &
\rev{ 0.00436 } &
\rev{ 0.00416 } &
\rev{ 0.00398 } &
\rev{ 0.00381 } &
\rev{ 0.00365 } &
\rev{ 0.00351 }\\
\textbf{MPMC} & 
\rev{ 0.00488 } &
\rev{ 0.00476 } &
\rev{ 0.00454 } &
\rev{ 0.00437 } &
\rev{ 0.00416 } &
\rev{ 0.00397 } &
\rev{ 0.00376 } &
\rev{ 0.00360 } &
\rev{ 0.00356 } &
\rev{ 0.00319 } &
\rev{ 0.00321 } &
\rev{ 0.00308 } &
\rev{ 0.00303 } &\\

\bottomrule
\end{tabular}
}
\end{table}

\subsection{On the role of the radius in the nearest neighbor graph}
\begin{figure}[ht!]
\vspace{-0.2cm}
\begin{minipage}[t]{0.48\textwidth}
\includegraphics[width=1.1\textwidth]{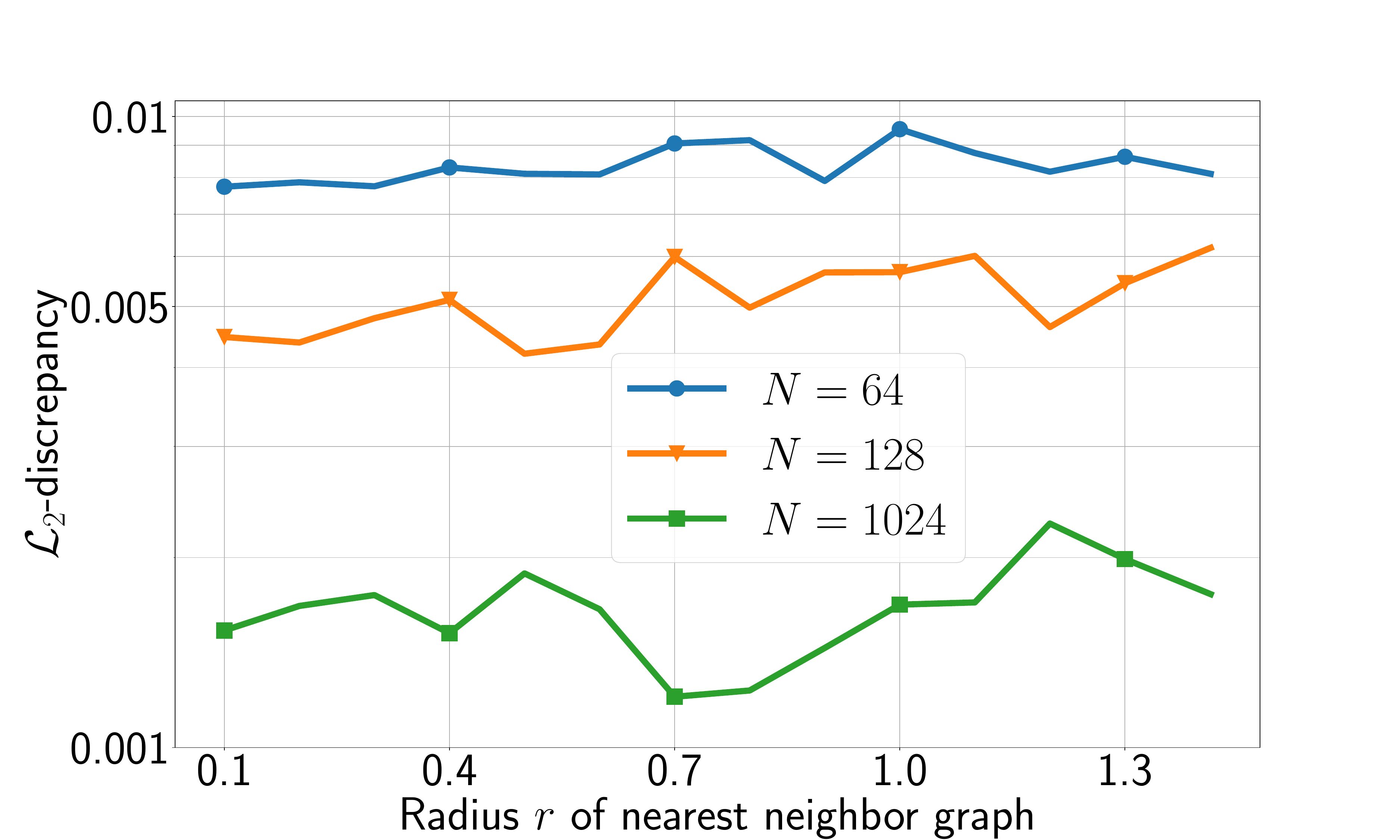}
\caption{$\cL_2$-discrepancy of MPMC points for increasing values of the radius of the underlying nearest neighbor computational graph ranging from $0.1$ to $\rev{\sqrt{2}}$ for different number of points $N=64,128,1024$ \rev{in $d=2$}.}
\label{fig:radius_ablation}    
\end{minipage}
\hfill
\begin{minipage}[t]{0.48\textwidth}
\includegraphics[width=1.1\textwidth]{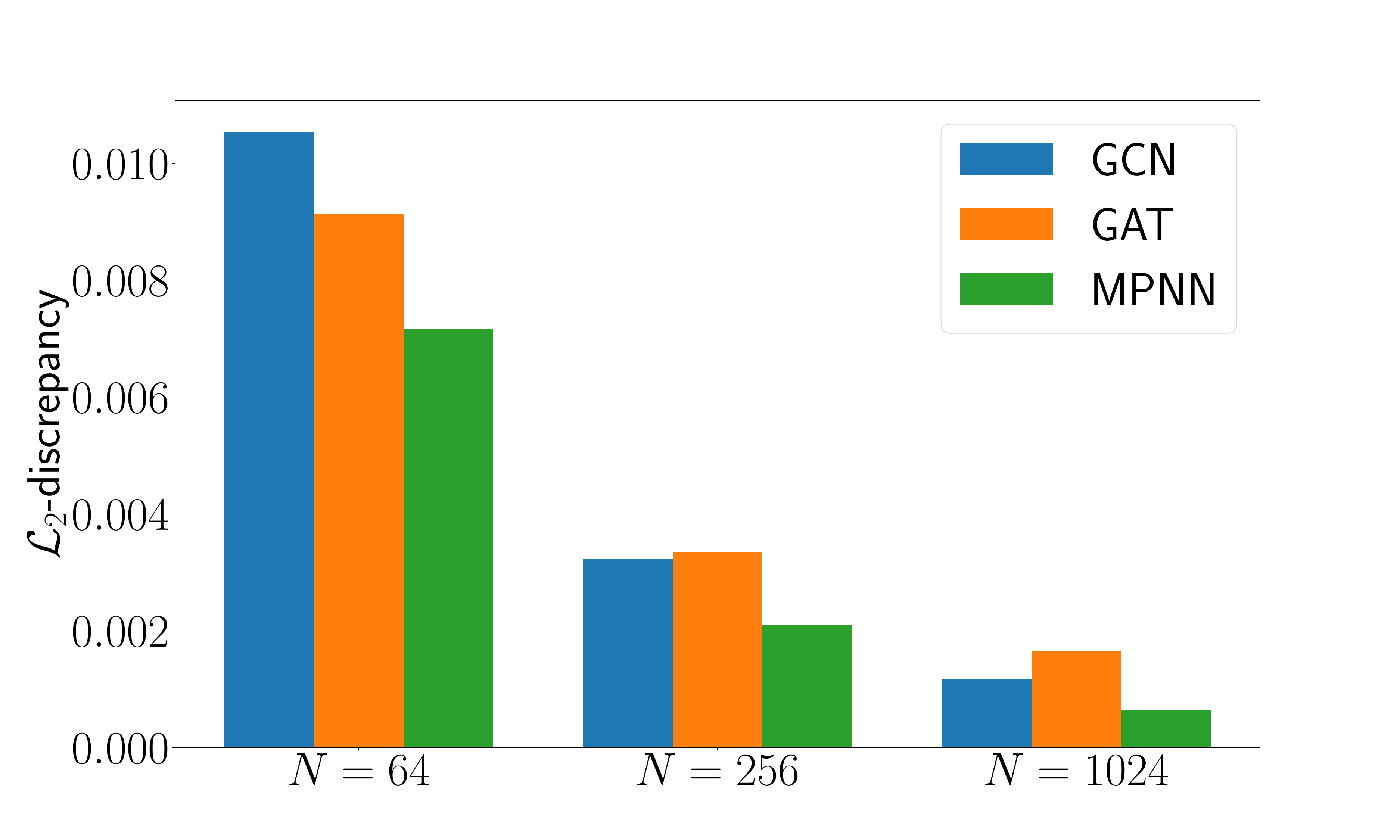}
\caption{$\cL_2$-discrepancy of MPMC for different choices of GNN architectures, i.e., GCN \cite{gcn}, GAT \cite{gat}, and MPNN \cite{mpnn} for three different number of points $N=64,256,1024$ \rev{in $d=2$}.}
\label{fig:model_ablation}    
\end{minipage}
\end{figure}

We recall from the main text, that our proposed MPMC method is based on GNNs that leverage $r$-radius nearest neighbors as the underlying computational graph, connecting nodes within a given radius $r$. How does the performance of MPMC depend on the radius $r$? Moreover, is it necessary to use GNNs? To answer this, we train $10$ MPMC models for different radius values $r=0.1,\dots,\rev{\sqrt{2}}$ for different number of points $N$ \rev{in $d=2$} and plot the resulting average $\cL_2$-discrepancy in \fref{fig:radius_ablation}. We can see that there is no correlation between the performance and a fixed radius $r$. Moreover, the performance appears to be not overly sensitive with respect to different values for the radius. Nevertheless, small variations of the performance with respect to the radius $r$ can be seen and it is thus advisable to include the radius to the set of tune-able hyperparameters of the model.

We further note that a radius of $r=0$ corresponds to zero edges in the underlying computational graph. Thus, the model becomes a deepset \cite{zaheer2017deep}, processing each point in the set individually without aggregating any neighborhood information. The average $\cL_2$-discrepancy of this deepset is approximately $0.073$ for $N=64$, $0.058$ for $N=128$, and $0.063$ for $N=1024$, \emph{i.e., between $9$ to over $40$ times worse than GNNs with $r\geq0.1$}. Moreover, the deepset fails to decrease the $\cL_2$-discrepancy for increasing number of points $N$. This highlights the necessity of using GNNs that aggregate geometric information from neighboring points for successfully generating low-discrepancy point sets. 

\subsection{On the role of the GNN architecture}
While we base our proposed MPMC model on MPNNs \cite{mpnn}, any other GNN architecture could be used instead. Therefore, it is natural to ask how the choice of the GNN architecture influences the performance of MPMC. To answer this, we test three different configurations of MPMC: one based on MPNNs, one based on GCNs \cite{gcn}, and one based on GATs \cite{gat}. We train all three configurations for different number of points $N=64,256,1024$ \rev{in $d=2$}, and provide the results as a bar plot in \fref{fig:model_ablation}. We can see that while GCNs and GATs outperform each other depending on the chosen number of points, MPNNs consistently produce point sets with the lowest $\cL_2$-discrepancy among all three configurations for all number of points considered here.

\begin{figure}[h!]
\vspace{-0.2cm}
\begin{minipage}[t]{.48\textwidth}
\includegraphics[width=1.1\textwidth]{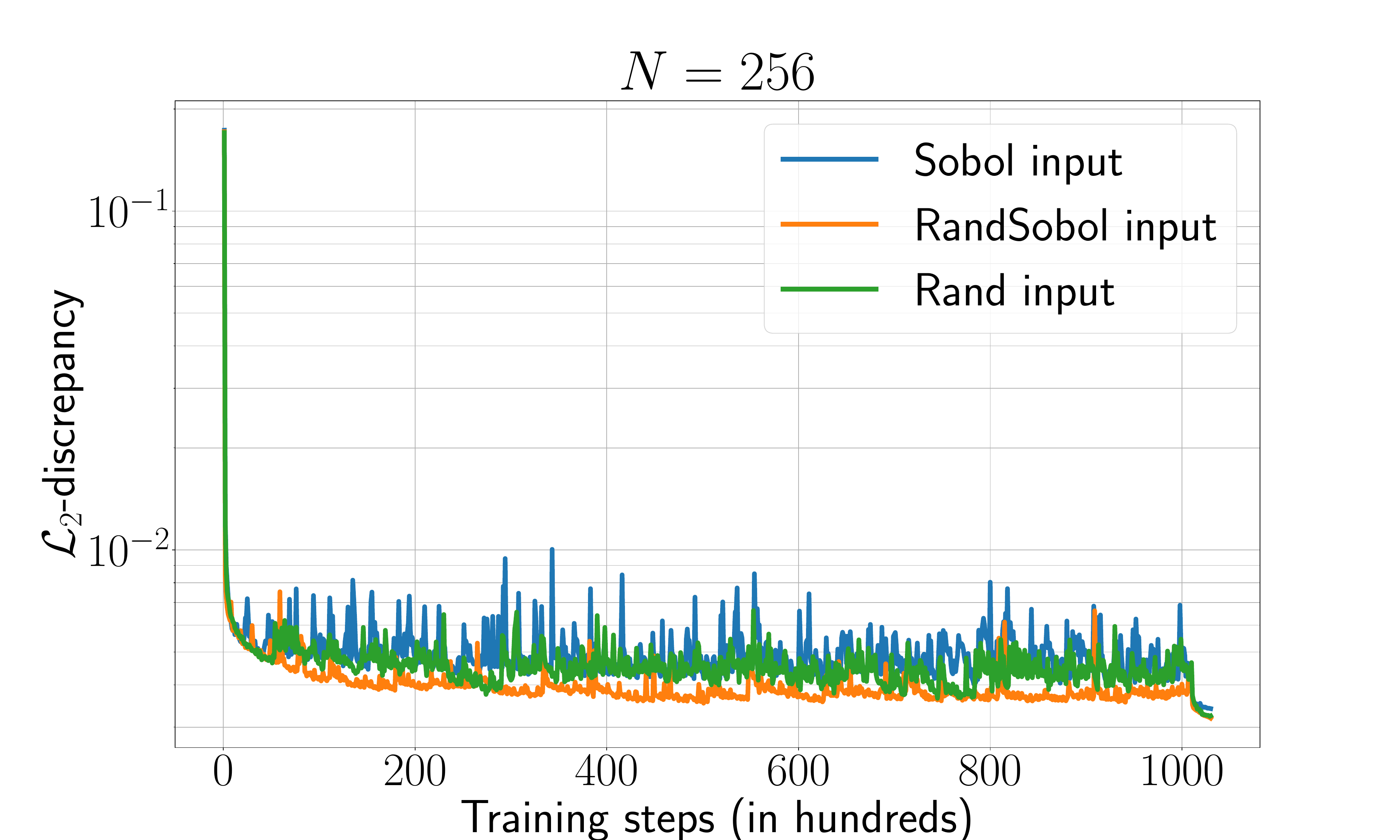}
\end{minipage}%
\hfill
\begin{minipage}[t]{.48\textwidth}
\includegraphics[width=1.1\textwidth]{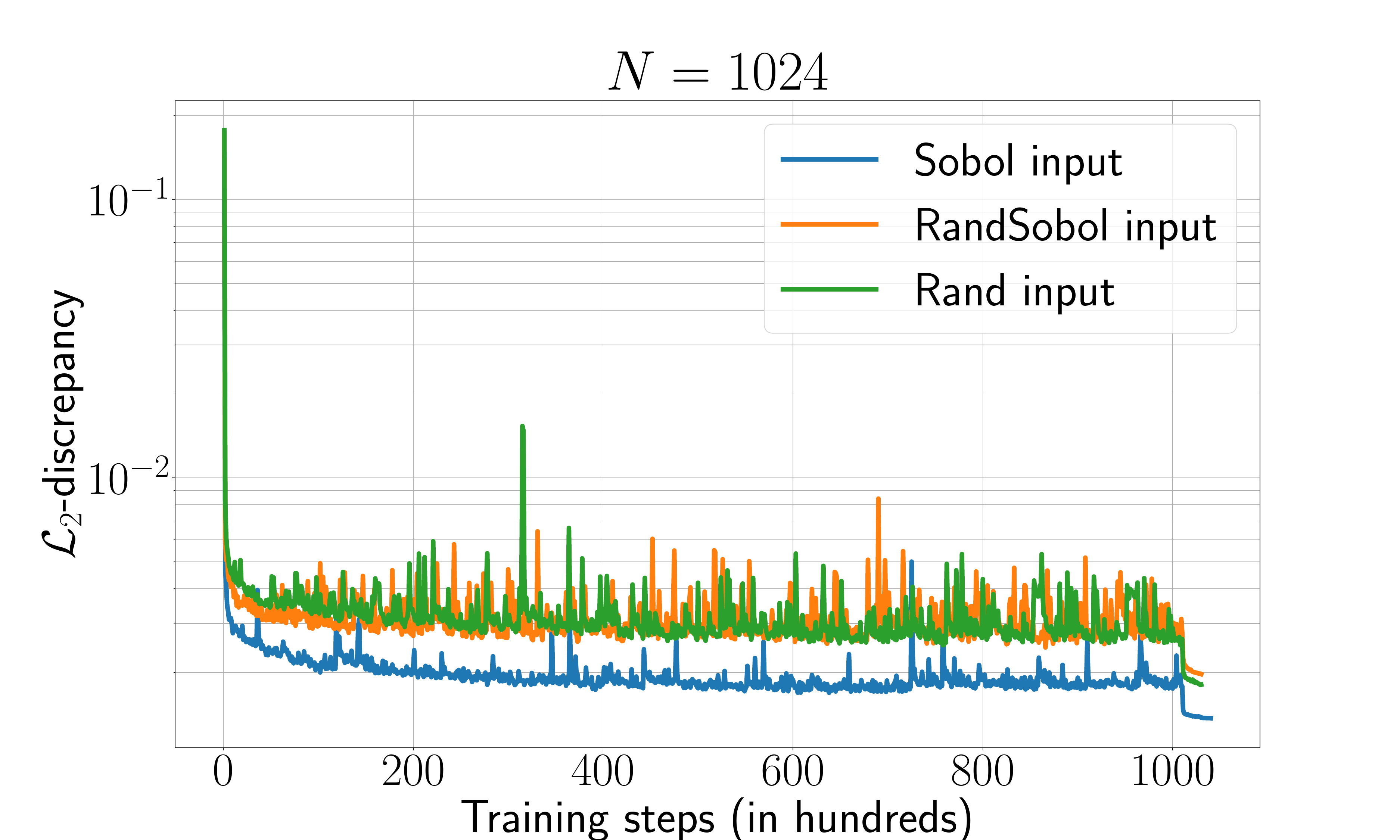}
\end{minipage}
\caption{$\cL_2$-discrepancy values of MPMC during training for three different types of input points, i.e., Sobol', Randomized Sobol', and random points, for $N=256$ and $N=1024$ \rev{in $d=2$}.}
\label{fig:input_type_ablation}
\vspace{-0.2cm}
\end{figure}

\subsection{On the role of the input points}
In the main text, we suggest three different input types to be transformed into low-discrepancy points via our MPMC framework, namely random points, Sobol', and randomized Sobol'. In this experiment, we empirically analyse how these different types influence the discrepancy of the resulting MPMC points. To this end, we train several MPMC models \rev{in $d=2$} based on the three different input types and report the average $\cL_2$-discrepancy during training for two different number of points $N=256,1024$ in \fref{fig:input_type_ablation}. We can see that on average either Sobol' or randomized Sobol' reach lower discrepancy values as well as exhibit faster convergence compared to random points. We note, however, that the single best MPMC model for each of the three different input point types yield almost identical discrepancy values. Thus, we conclude that Sobol' and randomized Sobol' points on average yield lower discrepancy values compared to random points, while at the same time the best performing input type has to be evaluated in practice for each number of points $N$.

\end{document}